\documentclass[12pt]{iopart}

\usepackage{amssymb}
\usepackage{multirow}
\usepackage{hyperref}

\usepackage{graphicx}
\usepackage{subfigure}
\usepackage{graphicx}

\usepackage{booktabs}
\usepackage{cite}
\expandafter\let\csname equation*\endcsname\relax
\expandafter\let\csname endequation*\endcsname\relax
\usepackage{booktabs}
\usepackage{makecell}

\hypersetup{
	colorlinks=true,
	linkcolor=red,   % 设置链接的颜色，可以根据需要调整
	citecolor=blue,   % 设置引用的颜色，可以根据需要调整
	filecolor=magenta,
	urlcolor=cyan,
	pdftitle={Your Title},   % 设置PDF文件的标题
	bookmarks=true,
	pdfpagemode=UseOutlines,
}

\begin{document}

\title[Hierarchical Multi-to-Single-Modal Knowledge Distillation for Disruption Prediction in EAST]{Hierarchical Multi-to-Single-Modal Knowledge Distillation for Disruption Prediction in EAST
}

\author{
    Qiang Chen\textsuperscript{1}, Xiao Wang\textsuperscript{1}, Hao Si\textsuperscript{1}, Qingquan Yang\textsuperscript{2*}, Meiwen Chen\textsuperscript{3}, Jianhua Yang\textsuperscript{2*}, Xiaofeng Han\textsuperscript{2}, Yunhu Jia\textsuperscript{2}, Ran Chen\textsuperscript{2}, Liang Wang\textsuperscript{2}, Jin Tang\textsuperscript{1}, Guosheng Xu\textsuperscript{2}
}

\address{\textsuperscript{1}School of Computer Science and Technology, Anhui University, Hefei 230031, China}
\address{\textsuperscript{2}Institute of Plasma Physics, Chinese Academy of Sciences, Hefei 230031, China}
\address{\textsuperscript{3}State Key Laboratory of Opto-Electronic Information Acquisition and Protection Technology, Anhui University, Hefei 230039, China}

\ead{\{yangqq, yangjh\}@ipp.ac.cn}
\vspace{10pt}

\begin{abstract}
% Plasma disruption is one of the most critical issues affecting the safe operation of tokamak devices.
Plasma disruption is a critical threat to tokamak safety.
Existing data-driven predictors mainly rely on time-series diagnostic signals, while visible images provide complementary spatial cues including plasma deformation, local brightening, and radiation-structure evolution.
Although the image modality improves the model's discriminative capability, it also substantially increases the computational cost during inference.
To address this issue, we propose a hierarchical multi-to-single-modal knowledge distillation framework for disruption prediction on a synchronized EAST multimodal dataset.
During training, visible images and time-series signals are used to train a multimodal teacher, which learns disruption precursor representations through Transformer-based encoders and a prototype-guided spatiotemporal hypergraph module.
During inference, only the time-series student is retained, with multimodal knowledge transferred through graph-structure-level, representation-level, and decision-level distillation.
On the 640-discharge EAST dataset, using the validation-selected thresholds with a 10 ms valid-warning constraint, the multimodal teacher achieves strong performance, with a TPR of 100.00\%, an FPR of 2.73\%, an F1 score of 96.00\%, and an AUC of 99.00\%.
The distilled unimodal student, which relies on time-series inputs during inference, still achieves a TPR of 91.66\%, an FPR of 2.73\%, an F1 score of 91.66\%, and an AUC of 97.88\%, while requiring only 3.75 ms for inference, 9.75G FLOPs, and 30.26M parameters.
Compared with the multimodal teacher, the distilled student achieves a $2.16\times$ inference speedup and reduces the FLOPs and parameter count by approximately 68.90\% and 47.85\%, respectively.
Experiments on the independent generalization dataset and the larger-scale dataset demonstrate the robustness and scalability of the proposed framework, while the similarity analysis confirms that the distilled student effectively inherits the teacher's knowledge in latent representation, decision boundary, and structural preference.
These results demonstrate that the proposed framework can preserve the discriminative advantages of multimodal learning while substantially reducing inference cost, and provide an effective route for efficient disruption prediction in EAST.
The source code of this paper will be released on \url{https://github.com/Event-AHU/OpenFusion}. 
\end{abstract}


\section{Introduction}

Plasma disruption is one of the major operational challenges in tokamak devices.
A disruption is accompanied by thermal quench, current quench, runaway electrons, and strong electromagnetic and thermal loads, all of which may cause substantial damage to the first wall, divertor, and other plasma-facing components~\cite{boozer2012theory,sugihara2007disruption}.
As future devices move toward larger size, higher plasma current, and higher stored energy, the operational risk associated with unmitigated disruptions will further increase.
Therefore, reliable disruption prediction capable of issuing alarms with sufficient lead time remains a prerequisite for disruption mitigation and active control~\cite{hender2007chapter}.

Existing data-driven disruption prediction studies mainly rely on time-series diagnostic signals and have been validated across a broad range of tokamaks, including JET~\cite{kates2019predicting,murari2020transfer,zheng2020disruption,vega2013results,aymerich2022disruption,pau2018first}, DIII-D~\cite{kates2019predicting,zhu2021hybrid}, ASDEX-U~\cite{aledda2015improvements}, C-Mod~\cite{zhu2021hybrid,rea2018disruption}, JT-60U~\cite{yoshino2003neural}, EAST~\cite{zhu2021hybrid,guo2023disruption,guo2021disruption}, J-TEXT~\cite{zheng2018hybrid,shen2023idp,zhong2021disruption}, 
HL-2A~\cite{yang2020disruption,yang2021depth}, HL-3~\cite{yang2025implementing}, KSTAR~\cite{kwon2021tokamak}, ADITYA-U~\cite{joshi2023analysis}, GOLEM~\cite{chandrasekaran2021data}, ST40~\cite{scarpari2024st40} and TCV~\cite{poels2025plasma}.
This demonstrates that disruption prediction based on time-series diagnostic signals has become the most mature and widely adopted paradigm in the field.
Within this dominant paradigm, existing disruption prediction studies based on time-series diagnostic signals have gradually expanded to several representative problem settings.
The first is high-performance single-device prediction, which focuses on improving predictive accuracy and ensuring sufficient warning time on a single tokamak~\cite{yang2020disruption,zheng2020disruption,lee2025machine,kim2024enhancing}.
For example, Guo et al. improved disruption prediction on EAST by training a full convolutional network on a large experimental database~\cite{guo2021disruption}, while Yang et al. enhanced the modeling of multichannel temporal precursors on HL-2A through a hybrid 1.5D CNN–LSTM framework~\cite{yang2020disruption}.
The second is cross-device transfer and few-shot deployment on new tokamaks, which aims to use knowledge learned from existing devices to support reliable prediction on target devices with limited data~\cite{shen2024cross,yang2025implementing}.
For example, Shen et al. applied domain adaptation to J-TEXT-to-EAST disruption prediction to reduce cross-device distribution discrepancy~\cite{shen2024cross}, and Yang et al. verified on HL-3 that deep learning models can still provide reliable warning under limited training data~\cite{yang2025implementing}.
The third is interpretability and plasma-state representation modeling, where the goal is not only to output a disruption probability, but also to explain prediction outcomes~\cite{poels2025plasma,liu2025interpretable,shen2023idp}.
More in detail, Poels et al. used a low-dimensional latent representation on TCV to characterize disruption risk and disruptivity~\cite{poels2025plasma}, while Liu et al. combined XGBoost and SHAP on EAST to reveal the contributions of key signals such as vertical displacement, $q_{95}$, and radiation~\cite{liu2025interpretable}.
The fourth is real-time deployment and application, which emphasizes whether prediction models can operate online and form a closed loop with the control system~\cite{lee2025real}. 
For example, Lee et al. integrated real-time disruption prediction, online plasma-control deployment, and MGI mitigation into a closed-loop framework on KSTAR, showing that rapid current ramp-down combined with MGI can provide a safer mitigation strategy for MA-level plasmas~\cite{lee2025real,rea2019real,hu2021real}.
Overall, disruption prediction based on time-series diagnostic signals has evolved from early single-device warning toward cross-device transfer, new-device deployment, real-time control, and interpretable modeling.
However, its information source is still largely limited to scalar signals and 1D profiles. 

Compared with conventional time-series diagnostic signals, visible images can directly provide spatial information such as plasma position, shape deformation, local brightening, and radiation-structure evolution, and thus may offer important complementary cues beyond conventional time-series diagnostic signals.
Recently, disruption-related studies based on visual information have begun to emerge.
Ratt\'a et al. and Spolladore et al. generated disruption-prevention-related alarms in JET using anomalous visual signatures captured by visible cameras~\cite{ratta2021phad,spolladore2023detection}. 
Kwon et al. achieved disruption event recognition in KSTAR using visible image sequences~\cite{kwon2021tokamak}. 
More recently, Chen et al. proposed Vi-DP on EAST and showed that pure video input also has substantial predictive capability~\cite{chen2026vi}.
These studies demonstrate that the video modality can provide important complementary discriminative information for disruption prediction.

Motivated by the above observations, multimodal fusion provides a natural direction for disruption prediction. 
By jointly exploiting the spatial information contained in visible images and the dynamical evolution information carried by time-series diagnostic signals, multimodal models have the potential to learn richer representations of disruption precursors. 
A few existing studies have provided preliminary evidence for this direction.
For example, Kim et al. combined video data with 0D parameters for multimodal disruption prediction on KSTAR, showing that the fusion of visual information and conventional diagnostics can improve both discriminative capability and warning performance~\cite{kim2024disruption}. 
However, studies on multimodal disruption prediction remain limited, and this direction still requires further investigation.

On the other hand, the introduction of the image modality also leads to a substantial increase in computational and inference cost. 
Kim et al. reported that the inference time of their multimodal model is approximately 60 ms~\cite{kim2024disruption}, while Vi-DP introduced a dedicated lightweight architecture to reduce inference time under the high-throughput setting of image inputs~\cite{chen2026vi}. 
Therefore, the key challenge is how to retain the discriminative advantages of multimodal learning while reducing the additional cost introduced by the image modality. 
To address this issue, we propose a multimodal knowledge distillation framework for disruption prediction on the EAST tokamak. 
During training, synchronized visible images and time-series diagnostic signals are used to construct a multimodal teacher model, which learns richer representations of disruption precursors. During inference, only a time-series-based student model is retained, thereby reducing the computational cost introduced by the image modality. 
To effectively transfer the multimodal knowledge from the teacher to the student, a multi-level distillation strategy is adopted. 
Graph-structure-level distillation transfers the higher-order relational patterns learned by the teacher in the spatiotemporal hypergraph, representation-level distillation constrains the student to learn the teacher’s intermediate representations, and decision-level distillation transfers the teacher’s prediction distribution and decision boundary.
In this way, the student model can inherit the discriminative capability of the multimodal teacher as much as possible while relying only on time-series diagnostic signals during inference.

The main contributions of this work are summarized as follows.
\begin{itemize}
    \item A multimodal-training but unimodal-inference framework is proposed for tokamak disruption prediction, which exploits the complementarity between visible images and time-series diagnostic signals while preserving the efficiency of time-series-based inference.
    \item A multimodal teacher model with integrated spatiotemporal hypergraphs is designed to explicitly capture higher-order correlations of disruption precursors in temporal evolution, spatial structure, and cross-modal semantics.
    \item A multi-level distillation strategy, including graph-structure-level, representation-level, and decision-level knowledge transfer, is designed to progressively transfer multimodal knowledge from the teacher to the time-series-based student.
\end{itemize}

The remainder of this paper is organized as follows. 
Section \ref{sec:Datasets} introduces the multimodal disruption dataset collected on EAST.
Section \ref{sec:design} presents the multimodal teacher model together with the multi-level knowledge distillation framework.
Section \ref{section:Experimental Setup} reports the experimental results, including performance comparison, efficiency evaluation, generalization analysis, larger-scale dataset evaluation, and visualization-based interpretability analysis.
Finally, Section \ref{sec:Conclusions} concludes the paper.

\section{EAST multimodal disruption dataset}
\label{sec:Datasets}

% \begin{figure}[h]
% 	\subfigure{
% 		\includegraphics[width=0.48\textwidth]{figures/time_delay.png}
% 	}
% 	\subfigure{
% 		\includegraphics[width=0.48\textwidth]{figures/time_dealy18.png}
% 		}
%     \caption{Temporal alignment of visible images and time-series diagnostic signals in the EAST multimodal dataset: (a) overall synchronization pipeline; (b) temporal delay distribution.}
% 	\label{fig:time_delay}
% \end{figure}

\begin{figure}[h]
    \centering
    \includegraphics[width=0.85\textwidth]{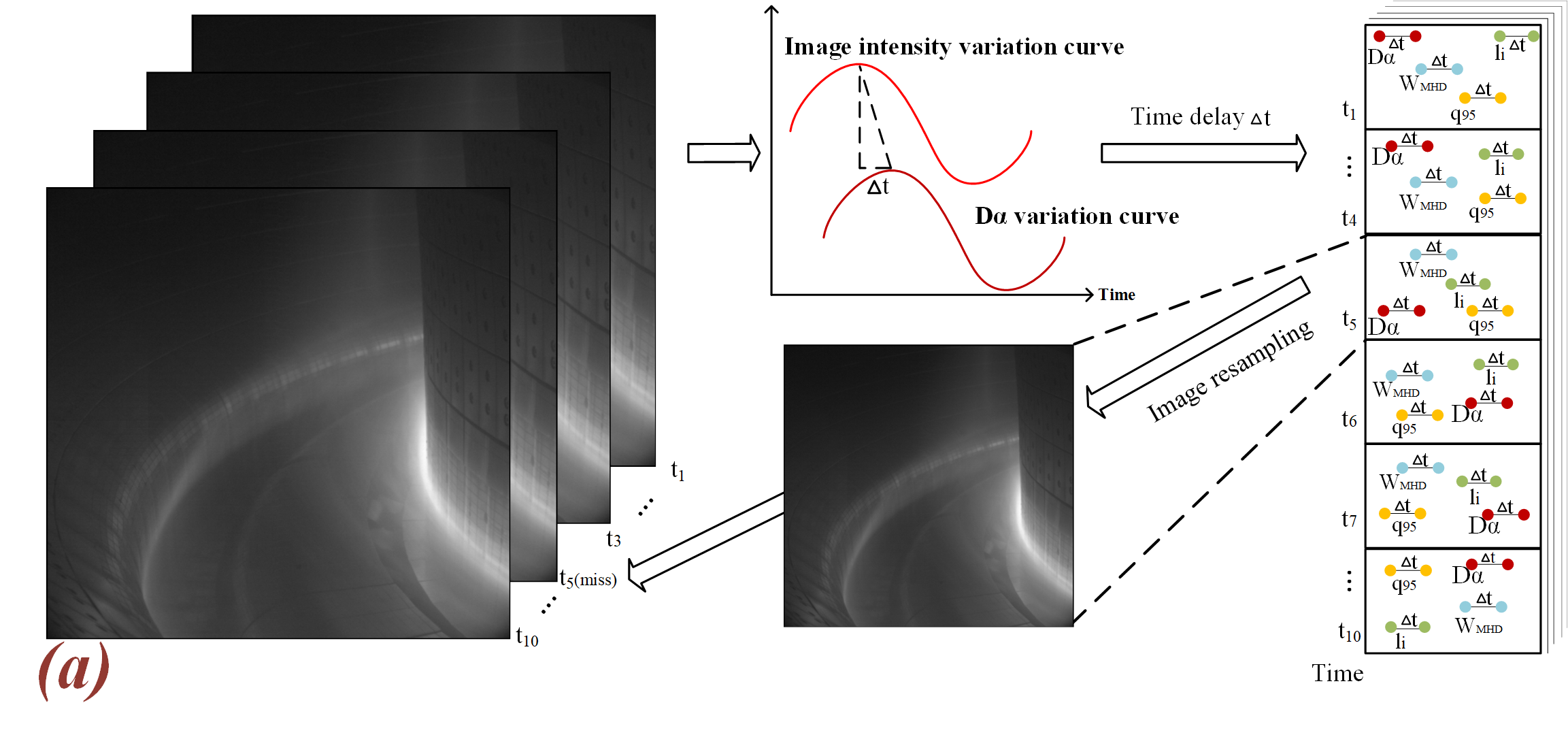}

    \vspace{-0.5em}

    \includegraphics[width=0.85\textwidth]{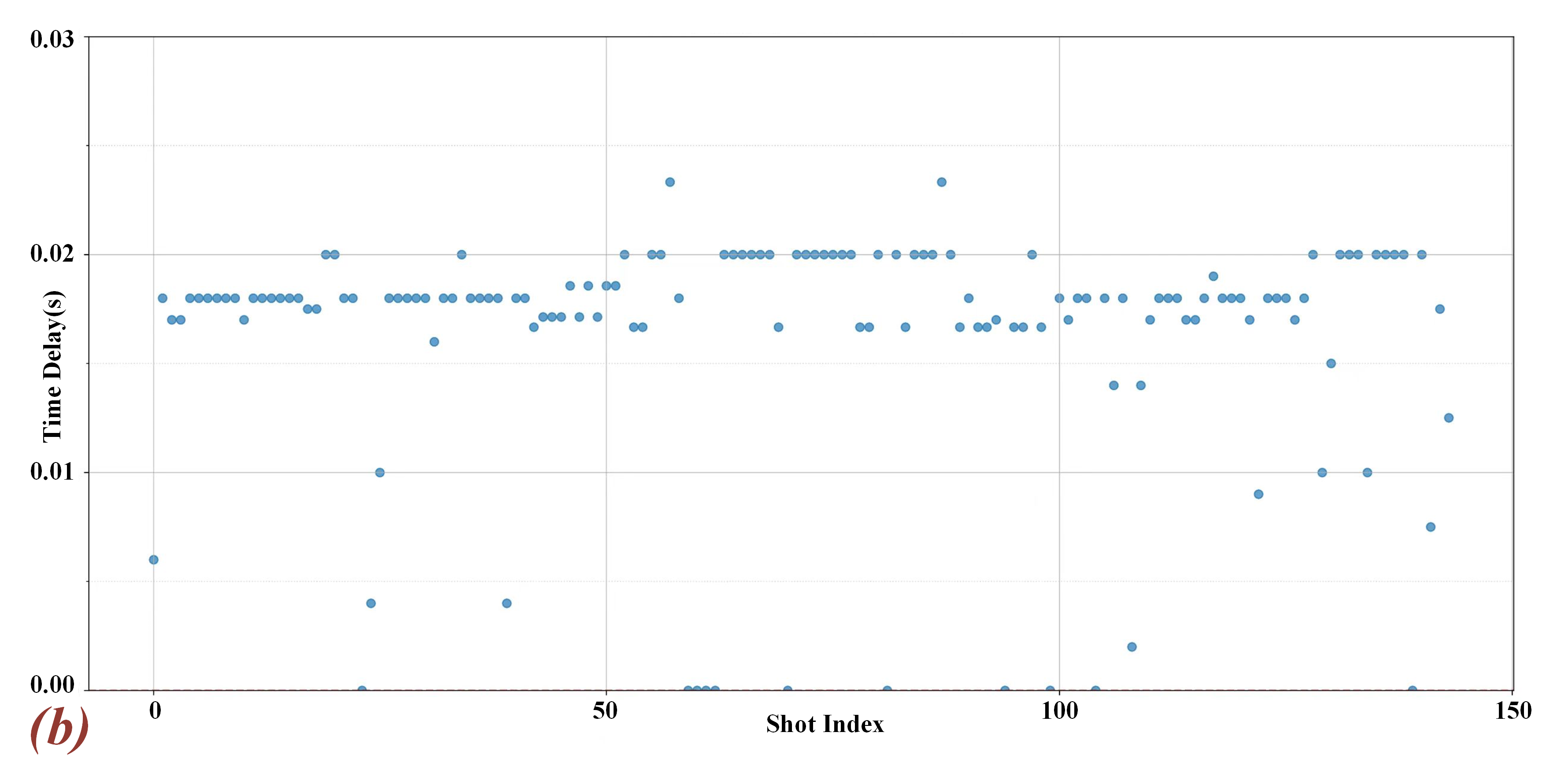}

    \vspace{-0.5em}

    \caption{Temporal alignment of visible images and time-series diagnostic signals in the EAST multimodal dataset: (a) overall synchronization pipeline; (b) temporal delay distribution.}
    \label{fig:time_delay}
\end{figure}

\begin{figure*}[t]
    \centering
    \includegraphics[width=\textwidth]{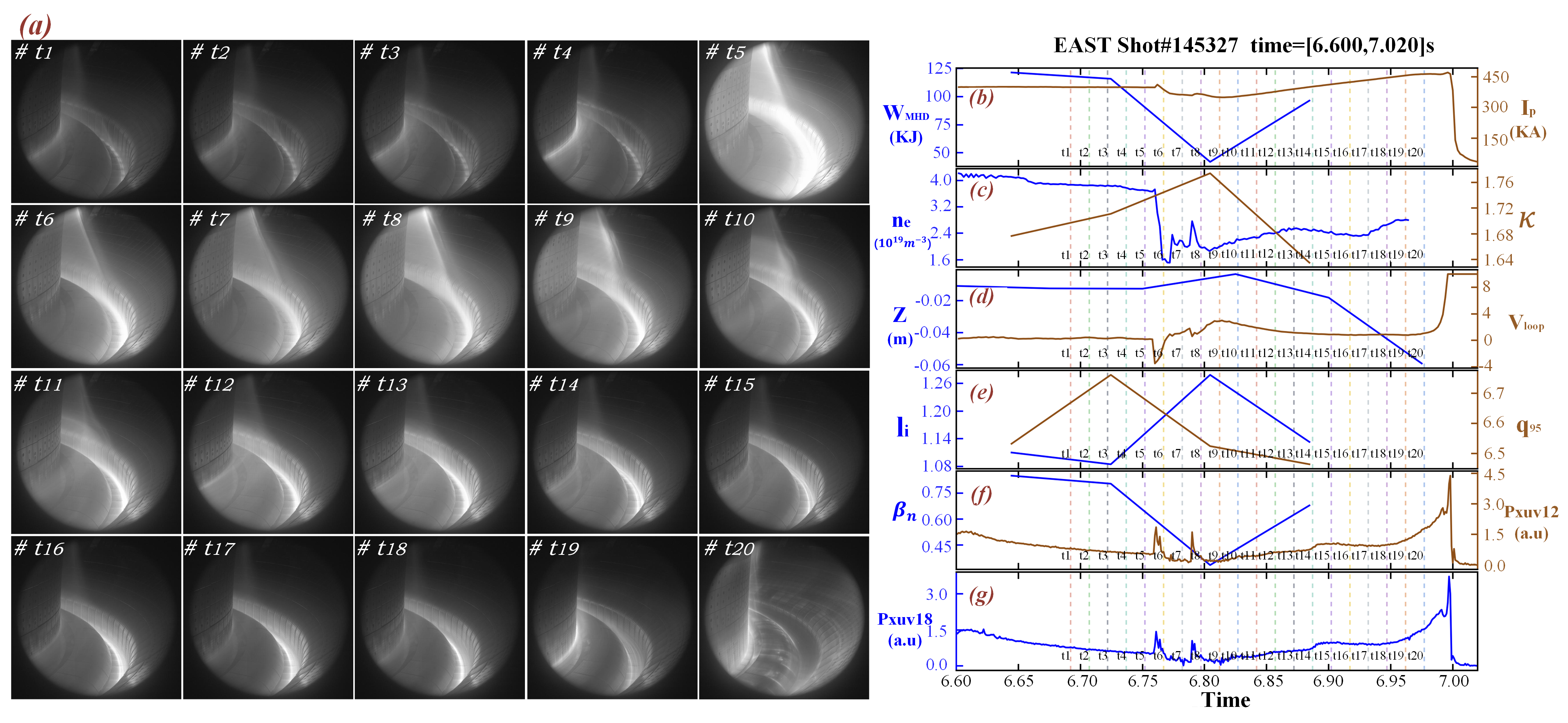}
    \caption{Representative disruption evolution of EAST shot \#145327: 
    (a) visible image sequence; 
    (b) evolution of plasma stored energy $W_{\mathrm{MHD}}$ and plasma current $I_p$; 
    (c) variations of plasma density $n_e$ and elongation ratio $\kappa$; 
    (d) variations of vertical displacement $Z$ and loop voltage $V_{\mathrm{loop}}$; 
    (e) variations of internal inductance $l_i$ and safety factor $q_{95}$; 
    (f) variations of normalized beta $\beta_n$ and edge radiation signal $P_{\mathrm{xuv12}}$; 
    (g) variation of core radiation signal $P_{\mathrm{xuv18}}$.}
    \label{fig:vis_dataset}
\end{figure*}

Due to differences in the triggering time among different diagnostic systems, a temporal delay may exist between the image data and the time-series diagnostic signals.
As shown in Figure~\ref{fig:time_delay}(a), to ensure temporal alignment of the multimodal data, the global brightness variation of the visible images is first converted into a temporal brightness curve, which is then cross-correlated with the time-series D$\alpha$ signal.
As shown in Figure~\ref{fig:time_delay}(b), the image modality exhibits an approximately \(18~\mathrm{ms}\) delay relative to the time-series diagnostic signals. 
Therefore, this delay is compensated for during data preprocessing, and both the image data and time-series diagnostic signals are uniformly resampled to a common \(1~\mathrm{kHz}\) time base.

Figure~\ref{fig:vis_dataset} illustrates the disruption evolution process of EAST Shot~\#145327.
Figure~\ref{fig:vis_dataset}(a) shows the visible images, while Figure~\ref{fig:vis_dataset}(b)--\ref{fig:vis_dataset}(g) present the corresponding time-series diagnostic signals.
It can be observed that, starting from \(t_4\), the vertical position \(Z\) of the plasma current centroid gradually moves upward. 
Subsequently, around \(t_{10}\), the \(Z\) signal gradually shifts downward, while a distinct radiation band appears in the lower region of the image, indicating that the plasma enters an unstable evolution stage and eventually terminates in a disruption. 
This example demonstrates that the evolution of the plasma state can be jointly characterized by visible images and time-series diagnostic signals.

To focus on the most damaging disruption events, only disruptive shots occurring during the flat-top phase are included in this study. 
The resulting dataset contains synchronized visible images and 11 time-series diagnostic signals, namely $I_p$, $Z$, $V_{\mathrm{loop}}$, $\kappa$, $l_i$, $q_{95}$, $\beta_n$, Pxuv12, Pxuv18, $W_{\mathrm{mhd}}$, and $n_e$. 
These signals cover plasma current, position, loop voltage, equilibrium shape, magnetohydrodynamic stability, radiation, and global plasma-state evolution. 
Their physical meanings and units are summarized in Table~\ref{tab:signals}.
% Depending on the camera setting, the image sampling rate ranges from approximately 100 to 2000~Hz, with an exposure time of about 2~$\mu$s. 
To standardize the model input, the images are padded on both sides and resized to $224\times224$.
% To ensure strict multimodal alignment, all samples are synchronized to a common 1~kHz time base and then segmented accordingly. 
Only shots with complete image data, time-series diagnostic signals, and disruption-time annotations are retained.

As shown in Table~\ref{tab:mm_dataset}, the dataset contains 640 discharges in total, including 447 training discharges, 96 validation discharges, and 97 test discharges. 
Among these, the number of disruptive shots in the training, validation, and test sets is 111, 24, and 24, respectively. 
With a 50~ms time window, the numbers of samples in the training, validation, and test sets are 107472, 88485, and 84137. 
In the training set, 79500 windows are non-disruptive and 27972 are disruptive windows. In the validation set, the corresponding numbers are 68231 and 20254; and in the test set, they are 69570 and 14567.
% Unless otherwise specified, this 640-discharge dataset is used for model training, validation, and primary testing. An additional non-overlapping EAST multimodal generalization dataset is introduced in Section~4.4 only for robustness evaluation.

\begin{table}[t]
\centering
\caption{Time-series diagnostic signals used in the multimodal disruption dataset.}
\label{tab:signals}
\begin{tabular}{lll}
\hline
Signal & Physical meaning & Unit \\
\hline
$I_p$ & Plasma current & kA \\
$Z$ & Plasma current centroid vertical position & m \\
$V_{\mathrm{loop}}$ & Loop voltage & V \\
$\kappa$ & Plasma elongation ratio & -- \\
$l_i$ & Plasma inductance & -- \\
$q_{95}$ & Safety factor & -- \\
$\beta_n$ & Normalized plasma pressure & -- \\
Pxuv12 & Radiation at the edge of the plasma & a.u. \\
Pxuv18 & Radiation at the core of the plasma & a.u. \\
$W_{\mathrm{MHD}}$ & Energy stored in plasma & kJ \\
$n_e$ & Plasma density & $10^{19}\,\mathrm{m}^{-3}$ \\
\hline
\end{tabular}
\end{table}

All samples are constructed using a sliding time window of length 50~ms. 
For non-disruptive shots in the training set, sample windows are generated within the valid analysis interval using a 50~ms window with a stride of 50~ms, and all windows are assigned a label of 0.
For disruptive shots in the training set, windows that lie entirely before $t_{\mathrm{dis}}-300$~ms are generated in the same manner as those of non-disruptive shots.
Within the final 300~ms danger zone of a disruptive shot, namely $[t_{\mathrm{dis}}-300,t_{\mathrm{dis}}]$, dense scanning is performed with a 50~ms window and a stride of 1~ms to characterize the continuous evolution of disruption precursors more precisely.

To avoid reducing the disruption evolution process to a coarse binary label, continuous soft labels are adopted. For any time instant $\tau$ within the danger zone, let
\begin{equation}
y(\tau)=
\left\{
\begin{array}{l l}
0,  & \Delta(\tau)>200\ \mathrm{ms},\\[4pt]
\frac{1}{1+\exp\!\left(-\frac{\tau-(t_{\mathrm{dis}}-100)}{15}\right)}, & 10<\Delta(\tau)\leq 200\ \mathrm{ms},\\[10pt]
1,  & \Delta(\tau)\leq 10\ \mathrm{ms}. 
\end{array}
\right.
\label{eq:soft_label_point}
\end{equation}
The label of a sample window starting at time $t$ is defined as the average of the millisecond-level labels inside the window:
\begin{equation}
Y(t)=\frac{1}{L}\sum_{\tau=t}^{t+L-1} y(\tau), \qquad L=50\ \mathrm{ms}.
\label{eq:soft_label_window}
\end{equation}
Compared with a simple hard label, this formulation better matches the continuous evolution of a disruption from distant precursors to imminent instability: the label remains 0 far from the disruption, saturates to 1 in the last 10~ms before the disruption, and rises smoothly in the 10--200~ms interval through a logistic transition.
For the validation and test sets, the full discharge is scanned using a 50~ms window and a 10~ms stride to mimic the online warning process. 
% The corresponding evaluation metrics and threshold-selection strategy are introduced in the experimental section.

\begin{table}[t]
\centering
\caption{Split of the EAST multimodal disruption dataset.}
\label{tab:mm_dataset}
\begin{tabular}{lcccc}
\hline
Discharge Type& Training & Validation & Test & Total \\
\hline
Non-disruption     & 336 & 72 & 73 & 481 \\
Disruption & 111 & 24 & 24 & 159 \\
All & 447 & 96 & 97 & 640 \\
\hline
\end{tabular}
\end{table}

\section{Model design}
\label{sec:design}

\subsection{Overall framework}
Tokamak disruption precursors are reflected in both the progressive deviation of time-series diagnostic signals and the spatial structural variation observed in visible images. Effective disruption prediction therefore requires not only the characterization of plasma-state evolution over time, but also the extraction of disruption-relevant spatial discriminative cues.

\begin{figure*}[t]
    \centering
    \includegraphics[width=\textwidth]{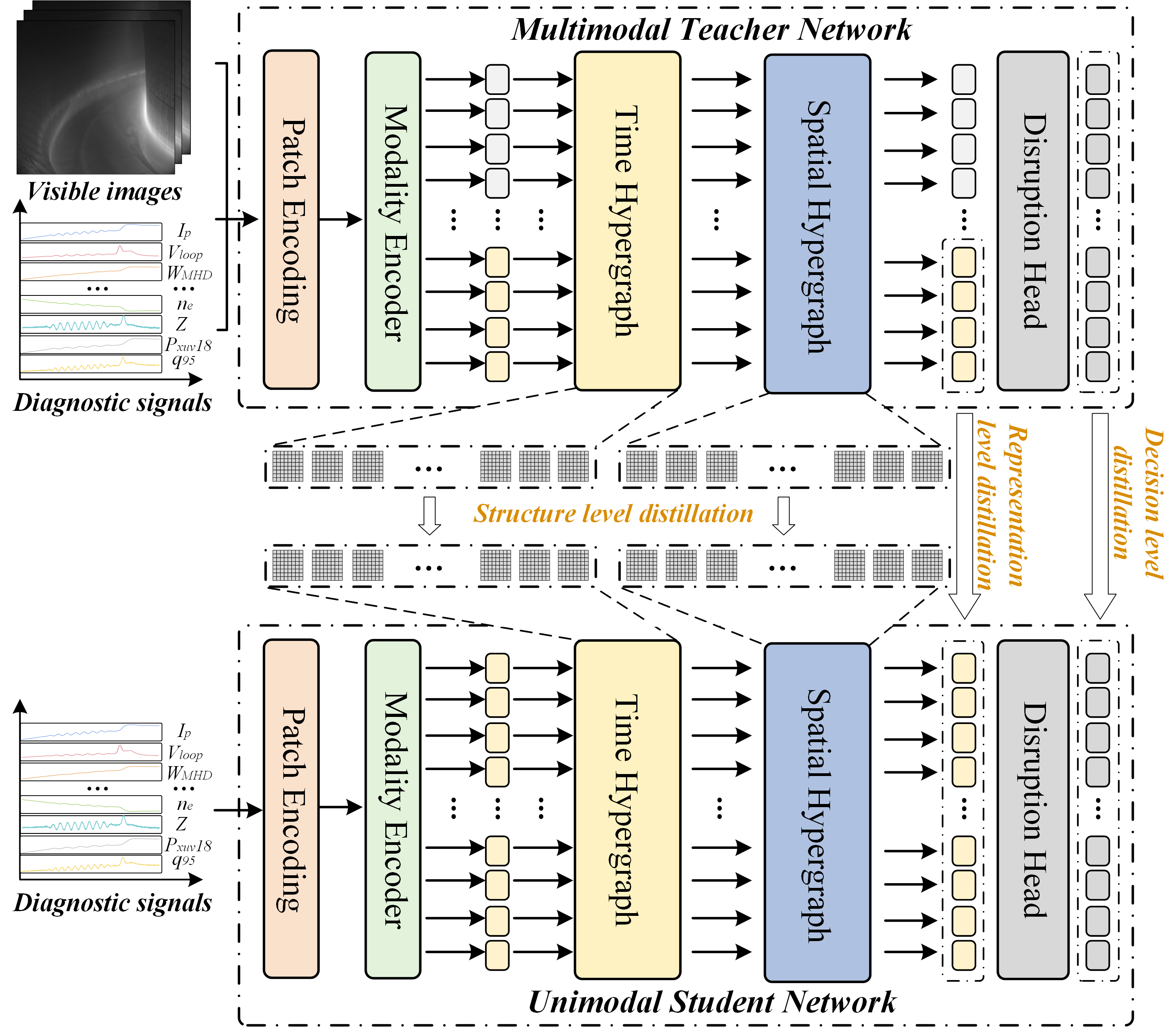}
    % \caption{Overview of the multimodal teacher–unimodal student framework and hierarchical knowledge distillation for disruption prediction in EAST.}
    \caption{Multimodal teacher--unimodal student framework and hierarchical knowledge distillation process for EAST disruption prediction. 
    The upper part shows the multimodal teacher model, while the lower part shows the time-series student model. 
    The student inherits the multimodal discriminative knowledge of the teacher through representation-level, decision-level, and structure-level distillation.}
    
    \label{fig:framework}
\end{figure*}

However, multimodal modeling does not simply amount to concatenating two types of features. 
For the image modality, disruption-related visual cues usually exhibit global dependencies across spatial regions and temporal evolution. 
For the time-series modality, the coordinated variation across temporal steps and signal channels likewise reflects the evolution of plasma states. 
A suitable encoder is therefore required to model long-range dependencies in a unified manner and to produce a consistent representation space. 
Conventional CNNs mainly focus on local image regions and are thus insufficient for capturing global visual structures. 
Although LSTM can model sequential dynamics through historical states, its unidirectional recurrent mechanism may lead to early-information attenuation in long sequences. 
By contrast, Transformer has been widely demonstrated to be effective for long-range dependency modeling. 
As illustrated in Figure~\ref{fig:framework}, in the encoding stage, Transformer is adopted for both the image and time-series modalities to obtain modality-specific representations with global contextual information.

At the fusion stage, the main issue is no longer the modeling of long-range dependencies within a single modality, but the characterization of higher-order precursor relations jointly formed by multiple temporal steps, multiple signal channels, and multiple spatial regions. 
The self-attention mechanism in a standard Transformer primarily models pairwise relations between tokens, which is not sufficient to fully describe such groupwise coupling patterns. 
Moreover, further stacking Transformer blocks would introduce additional computational cost. 
For this reason, a spatiotemporal hypergraph module is introduced at the fusion stage. 
By allowing each hyperedge to connect multiple nodes simultaneously, the spatiotemporal hypergraph can explicitly capture higher-order couplings across temporal steps, signal channels, and spatial regions, and thus provides a more natural representation of the groupwise relational patterns embedded in disruption precursors.

On the other hand, although multimodal models can learn richer disruption precursor information, the introduction of the image modality also substantially increases the computational burden during inference. 
As shown in Figure~\ref{fig:framework}, the multimodal model is not directly used as the final disruption predictor. 
Instead, a multimodal teacher--unimodal student framework is constructed. 
Through multi-level distillation, the teacher’s knowledge at the representation, decision, and graph-structure levels is progressively transferred to the student model, so that the student preserves inference efficiency while inheriting the discriminative advantages provided by multimodal learning.

The following subsections describe the modality encoders, the prototype and spatiotemporal hypergraph fusion module, and the multi-level distillation strategy in detail.

\begin{figure*}[t]
    \centering
    \includegraphics[width=\textwidth]{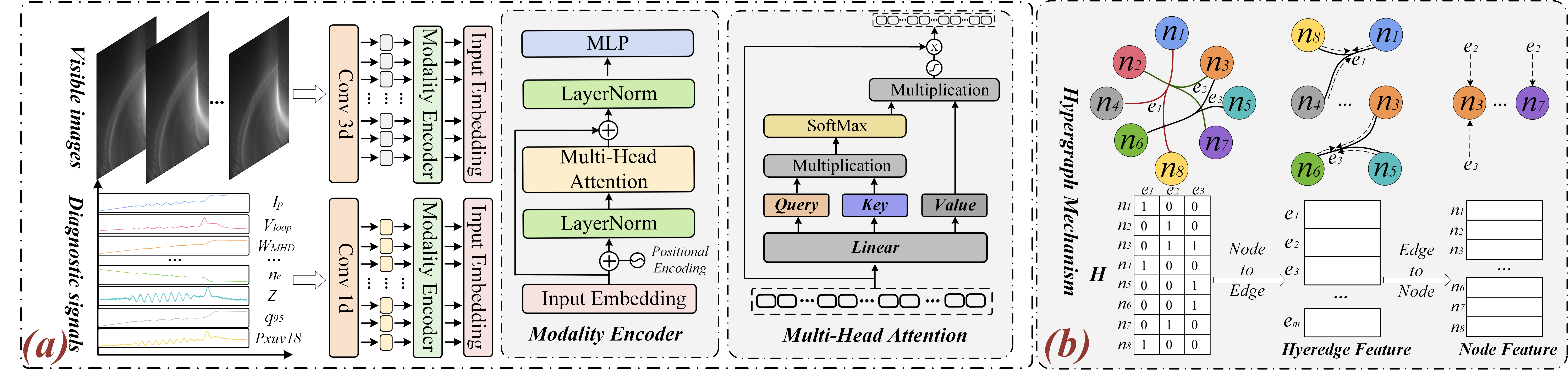}
    \caption{
    (a) Architecture of the modality encoders. Visible image sequences and diagnostic signals are mapped into tokens through 3D/1D convolutions and then fed into Transformer-based modality encoders to extract global features.
    (b) Hypergraph propagation mechanism. The encoded features are treated as nodes, while learnable prototypes serve as hyperedge anchors. Sparse node--hyperedge association matrices are constructed through similarity matching and top-$k$ selection, and are further used to propagate information from nodes to hyperedges and then back from hyperedges to nodes.}
    \label{fig:hypergraph1}
\end{figure*}

\subsection{Modality encoders for image and signal branches}
% \subsection{Unified modal representation learning}

Figure~\ref{fig:hypergraph1}(a) illustrates the modal representation learning component of the proposed model. 
In order to establish compatible latent representations across image and signal inputs, Transformer-style token-based encoders are adopted in both branches. 
Local spatiotemporal patches from image sequences and local temporal segments from diagnostic signals are both projected into token representations, thereby establishing a unified representational basis for subsequent multimodal alignment, higher-order relation modeling, and knowledge distillation.

Within the encoder, multi-head self-attention is used to model global token dependencies. Its expanded form can be written as
\begin{equation}
Q_h = XW_h^Q,\qquad
K_h = XW_h^K,\qquad
V_h = XW_h^V,
\end{equation}
\begin{equation}
\mathrm{head}_h
=
\mathrm{Softmax}\!\left(
\frac{Q_hK_h^\top}{\sqrt{d_h}}
\right)V_h,
\end{equation}
\begin{equation}
\mathrm{MHA}(X)
=
\mathrm{Concat}(\mathrm{head}_1,\ldots,\mathrm{head}_H)W^O,
\end{equation}
where \(H\) denotes the number of attention heads and \(d_h\) is the channel dimension of each head. 
The encoder block is updated as
\begin{equation}
\hat{X}=X+\mathrm{MHA}(\mathrm{LN}(X)),
\end{equation}
\begin{equation}
X^{\mathrm{out}}=\hat{X}+\mathrm{MLP}(\mathrm{LN}(\hat{X})).
\end{equation}
% Multi-head attention enables the model to capture long-range dependencies across time, regions, and channels.

For the image branch, each input window contains 12 visible frames. 
A 3D convolution is first used to embed local spatiotemporal blocks from consecutive frames, mapping the raw image sequence into spatiotemporal tokens. 
Then, the tokens are fed into the modality encoder for feature extraction.

For the signal branch, the input is an \(11\times 50\) temporal window, where 11 denotes the number of diagnostic channels and 50 denotes the number of sampled points within the 50~ms interval. 
A 1D convolution is first used to map local temporal segments in each channel into tokens. 
The token sequence is subsequently processed by the modality encoder to model the initial relations among temporal segments and diagnostic variables.

\begin{figure*}[t]
    \centering
    \includegraphics[width=\textwidth]{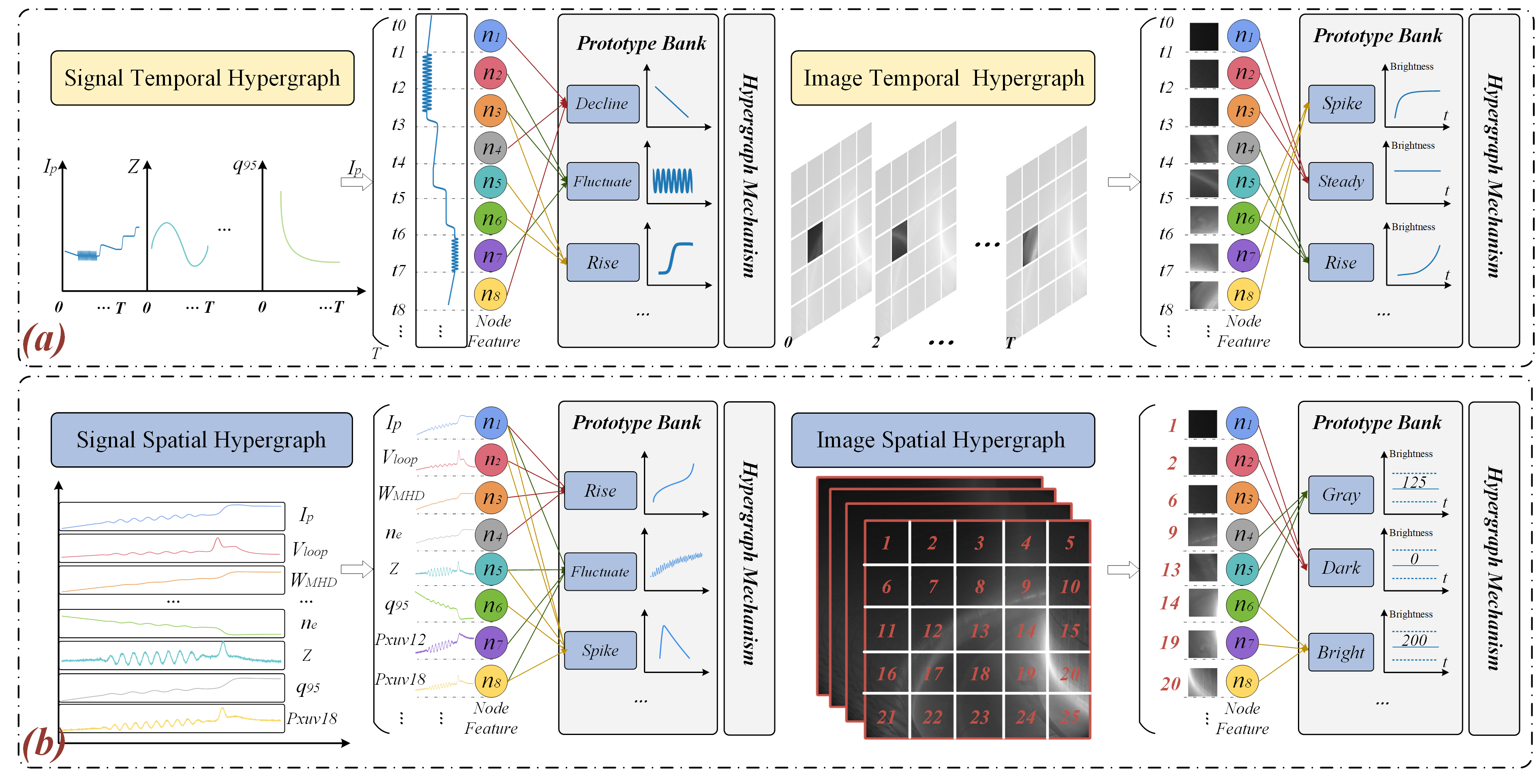}
    \caption{Prototype-guided spatiotemporal hypergraph construction in the diagnostic signal and image branches. 
    (a) Temporal hypergraphs are constructed along the temporal dimensions of diagnostic signals and image patches to model the temporal evolution of each diagnostic variable or fixed spatial region.
    (b) Spatial hypergraphs are constructed along the diagnostic-variable and image-patch dimensions to capture correlations among different channels or spatial regions at the same time instant.}
    \label{fig:hypergraph2}
\end{figure*}
% \begin{figure*}[t]
%     \centering
%     \includegraphics[width=\textwidth]{figures/hy_graph.png}
%     \caption{Architecture of the modality encoders and the prototype-guided spatiotemporal hypergraph module.}
%     \label{fig:hypergraph}
% \end{figure*}

\subsection{Spatiotemporal hypergraph learning with prototype-guided hyperedges}
% \subsection{Spatiotemporal hypergraph learning with learnable edge anchors}
Token representations produced by the modality encoders are still insufficient to fully characterize the higher-order dependencies embedded in disruption precursors. 
For disruption prediction, the key information is often embodied in group-wise evolution patterns jointly formed by multiple temporal segments, multiple diagnostic channels, or multiple local regions.

Ordinary graphs~\cite{zhang2019graph} mainly describe pairwise relations between nodes, whereas a hypergraph allows one hyperedge to connect multiple nodes simultaneously and is therefore more suitable for representing the higher-order organization of disruption precursors. 
For this reason, prototype-guided temporal and spatial hypergraphs are introduced in both the signal branch and the image branch to explicitly model higher-order relations across time and across nodes.

Figure~\ref{fig:hypergraph1}(b) illustrates the hypergraph mechanism adopted. 
Let the input node features be \(X=\{x_i\}_{i=1}^{M}\in\mathbb{R}^{M\times D}\), where \(M\) is the number of nodes and \(D\) is the feature dimension. 
To establish node--hyperedge associations, a set of \(K\) learnable edge-anchor representations is introduced, which are referred to as prototypes, denoted by \(P=\{p_k\}_{k=1}^{K}\in\mathbb{R}^{K\times D}\).
Then, the similarity score between node \(x_i\) and prototype \(p_k\) is computed as
\begin{equation}
S_{ik}=\frac{(W_qx_i)^\top(W_kp_k)}{\sqrt{D}},
\end{equation}
where \(W_q\) and \(W_k\) are learnable linear projection matrices that map node features and prototype features into the same matching space. 
The resulting score \(S_{ik}\) measures how strongly node \(x_i\) is associated with prototype \(p_k\). 
For each node, only the top-\(k\) prototypes with the highest scores are retained, and their normalized weights form a sparse association matrix \(H\).

Based on this association structure, hypergraph message passing proceeds through two successive steps, namely \emph{node-to-edge} and \emph{edge-to-node} propagation. 
First, nodes belonging to the same hyperedge are aggregated into a hyperedge feature:
\begin{equation}
e_k=\frac{\sum_i H_{ik}W_vx_i}{\sum_i H_{ik}+\varepsilon},
\end{equation}
where \(W_v\) is a learnable linear transformation matrix that maps node features into the hyperedge aggregation space before aggregation, and \(H_{ik}\) denotes the association weight between node \(i\) and hyperedge \(k\). 
This equation means that all nodes associated with hyperedge \(k\) are combined by a weighted average according to \(H_{ik}\), yielding the hyperedge feature \(e_k\). 
Second, the hyperedge features are propagated back to the nodes to update their representations:
\begin{equation}
\tilde{x}_i=\mathrm{LN}\!\left(W_vx_i+\mathrm{MLP}\!\left(\sum_k H_{ik}W_ee_k\right)\right),
\end{equation}
where \(W_e\) is another learnable linear transformation matrix that maps hyperedge features into the node update space. 
\(\sum_k H_{ik}W_ee_k\) represents the aggregated contextual information received by node \(i\) from all hyperedges connected to it, with \(H_{ik}\) controlling the contribution of hyperedge \(k\) to the update of node \(i\). 

This mechanism is applied to both the signal branch and the image branch, as shown in Figures~\ref{fig:hypergraph2}(a) and \ref{fig:hypergraph2}(b). 
In the signal branch, the output tokens of the modality encoder are rearranged into \([B,V,T,D]\), where \(V\) denotes the number of diagnostic variables and \(T\) denotes the temporal length of the window. 
The temporal hypergraph is constructed along the temporal dimension by reshaping the input into \([B\times V,T,D]\), so that the model learns the temporal evolution pattern of each variable within a window. 
The spatial hypergraph is then constructed along the variable dimension by reshaping the features into \([B\times T,V,D]\), in order to model the coupling among diagnostic channels at the same time instant. 
% The former corresponds to the left part of figure~\ref{fig:hypergraph}(c), whereas the latter corresponds to the left part of figure~\ref{fig:hypergraph}(d).

In the image branch, the encoded tokens are rearranged into \([B,N,F,D]\), where \(N\) denotes the number of spatial patches and \(F\) denotes the number of sampled frames. 
The temporal hypergraph is constructed along the frame dimension by reshaping the input into \([B\times N,F,D]\), which allows the model to capture the local evolution pattern of a fixed patch across consecutive frames. 
The spatial hypergraph is then constructed along the patch dimension by reshaping the input into \([B\times F,N,D]\), in order to model the coordinated variation among different spatial regions within the same frame. 
% The former corresponds to the right part of figure~\ref{fig:hypergraph}(c), whereas the latter corresponds to the right part of figure~\ref{fig:hypergraph}(d).

\subsection{Multi-level knowledge distillation}
\label{section:muti-level}
% Hierarchical knowledge transfer from the multimodal teacher to the signal student
Figure~\ref{fig:framework} illustrates the hierarchical knowledge-transfer process.
The difference between the multimodal teacher and the time-series-based student is not limited to the final prediction output.
It also lies in the intermediate representation and in the way higher-order structure is organized. 
If only intermediate features are aligned, the student may still fail to inherit the teacher's discriminative boundary and prototype organization. 
If the student is constrained only at the output level, the representational gain and structural inductive bias introduced by multimodal modeling cannot be fully transferred. 
For this reason, a hierarchical distillation strategy is adopted, consisting of representation-level, decision-level, and graph-structure-level knowledge transfer, corresponding to representation distillation, logit distillation, and graph distillation, respectively.
The objective is to transfer the signal-relevant knowledge of the multimodal teacher to the time-series-based student at the representation, decision, and structural levels.

\noindent\textbf{Graph-structure level.}
The graph-structure-level knowledge transfer corresponds to graph distillation, which transfers the sparse node-to-hyperedge association learned by the teacher in the temporal and spatial hypergraphs. 
Its purpose is to guide the student to inherit the teacher's signal-relevant higher-order relational patterns, including dependencies across temporal steps and correlations among diagnostic channels.

Let \(H^T\) and \(H^S\) denote the association matrices of the teacher and the student, respectively. 
To emphasize the most informative structural relations, the distillation constraint is imposed only on the top-\(k\) hyperedges retained by the teacher, so that the student focuses on learning the most discriminative higher-order association patterns provided by the teacher. 
Accordingly, the graph distillation loss is defined as
\begin{equation}
\mathcal{L}_{\mathrm{structure}}
=
-\frac{1}{N}\sum_{i=1}^{N}\sum_{k}
H_{ik}^{T}\log H_{ik}^{S},
\end{equation}
where \(H_{ik}\) denotes the association weight between node \(i\) and hyperedge \(k\), and the loss is computed only on the top-\(k\) hyperedges activated by the teacher.
This design is consistent with the sparse top-\(k\) incidence construction adopted in the proposed model.
Rather than forcing the student to reproduce the full structural distribution of the teacher, it encourages the student to inherit the most informative node-to-hyperedge association preference of the teacher. 
Since the temporal and spatial hypergraphs capture different aspects of higher-order dependency, this structural distillation constraint is imposed on both types of hypergraphs. 

\noindent\textbf{Representation level.}
The representation-level knowledge transfer corresponds to representation distillation, which constrains the intermediate representation of the student signal branch to approximate the multimodally enhanced signal feature of the teacher. 
Its purpose is to transfer the teacher's latent representation knowledge, enabling the student to learn more discriminative feature representations from time-series diagnostic signals.
Let $f^S$ and $f^T$ denote the student and teacher representations, respectively. The representation loss is defined as
\begin{equation}
L_{\mathrm{representation}} = \mathrm{SmoothL1}(f^S, f^T).
\end{equation}
Representation distillation further extends knowledge transfer to the latent space, enabling students to form more discriminative latent representations.
% Smooth L1 is preferred here over plain mean squared error because disruption prediction contains both relatively steady windows and highly transitional near-disruptive windows, which may induce local scale differences and feature outliers. Smooth L1 is more robust to such deviations while still encouraging effective feature alignment.

\noindent\textbf{Decision level.}
The decision-level knowledge transfer corresponds to logit distillation, which uses the teacher's logits as soft supervisory signals. 
Its purpose is to transfer the teacher's prediction distribution and soft decision boundary, enabling the student to inherit the teacher's fine-grained discriminative knowledge.
Let \(z^{T}\) and \(z^{S}\) denote the teacher and student logits, respectively. A temperature-scaled KL divergence is adopted:
\begin{equation}
\mathcal{L}_{\mathrm{decision}}
=
T^{2}\,
\mathrm{KL}\!\left(
\mathrm{Softmax}\!\left(\frac{z^{T}}{T}\right)
\;\middle\|\;
\mathrm{Softmax}\!\left(\frac{z^{S}}{T}\right)
\right).
\end{equation}
Logit distillation enables students to learn the teacher’s fine-grained judgments regarding sample uncertainty, thereby more accurately inheriting the decision boundaries formed by the multimodal teacher.

The overall objective of the student is therefore
\begin{equation}
\mathcal{L}
=
\mathcal{L}_{\mathrm{task}}
+
\lambda_{\mathrm{1}}
\left(
\mathcal{L}_{\mathrm{structure}}^{\mathrm{time}}
+
\mathcal{L}_{\mathrm{structure}}^{\mathrm{space}}
\right)
+
\lambda_{\mathrm{2}}\mathcal{L}_{\mathrm{representation}}
+
\lambda_{\mathrm{3}}\mathcal{L}_{\mathrm{decision}},
\end{equation}
where \(\mathcal{L}_{\mathrm{task}}\) denotes the supervised loss. 
In the current implementation, \(\lambda_{\mathrm{1}}=1\), \(\lambda_{\mathrm{2}}=0.9\), and \(\lambda_{\mathrm{3}}=0.1\).

\section{Experimental results and analysis}
\label{section:Experimental Setup}
\subsection{Experimental setup}

Experiments are conducted on an NVIDIA A100 GPU under Ubuntu 22.04.3 LTS with Python 3.10.18 and PyTorch 2.6.0~\cite{paszke2019pytorch}. 
The multimodal teacher network is trained with BCE loss with logits.
The distillation model is trained with the multi-level distillation loss defined in Section~\ref{section:muti-level} together with BCE loss with logits. 
All models are optimized with AdamW~\cite{loshchilov2019decoupled} for 10 epochs.

The decision threshold is selected on the validation set, and model performance is evaluated on complete discharge sequences from the test set using the selected threshold.
The optimal threshold corresponds to the highest $F_1$ score on the validation set. 
The $F_1$ score is defined as
\begin{equation}
F_1 = \frac{2 \times \mathrm{Precision} \times \mathrm{Recall}}{\mathrm{Precision} + \mathrm{Recall}},
\end{equation}
\begin{equation}
\mathrm{Precision} = \frac{TP}{TP + FP}, \qquad
\mathrm{Recall} = \frac{TP}{TP + FN}.
\end{equation}

For a disruptive shot, if the predicted probability first exceeds the threshold at time $t_{\mathrm{alarm}}$ and satisfies $t_{\mathrm{dis}}-t_{\mathrm{alarm}}\geq 10~\mathrm{ms}$, it is counted as a true positive (TP). 
Otherwise, it is counted as a false negative (FN).
For a non-disruptive shot, it is counted as a false positive (FP) if the predicted probability exceeds the threshold at any time.
Otherwise, it is counted as a true negative (TN). 
The true positive rate (TPR) is the proportion of disruptive shots correctly identified.
The false positive rate (FPR) is the proportion of non-disruptive shots incorrectly classified as disruptive.
These quantities are defined as
\begin{equation}
TPR = \frac{TP}{TP + FN}, \qquad
FPR = \frac{FP}{FP + TN}.
\end{equation}
The receiver operating characteristic (ROC) curve and the corresponding AUC are used to evaluate the overall discrimination capability under different thresholds.
The average warning time measures the early-warning capability of the model for disruptive discharges.
For the $i$-th correctly detected disruptive discharge, the warning time is defined as the time difference between the annotated disruption time and the first alarm time:
\begin{equation}
T_{\mathrm{alarm}}^{(i)} = t_{\mathrm{dis}}^{(i)} - t_{\mathrm{alarm}}^{(i)} .
\end{equation}
The average warning time is then obtained by averaging the warning times over all true-positive disruptive discharges:
\begin{equation}
\mathrm{Avg.Alarm} =
\frac{1}{N_{\mathrm{TP}}}
\sum_{i \in \mathcal{D}_{\mathrm{TP}}}
\left(t_{\mathrm{dis}}^{(i)} - t_{\mathrm{alarm}}^{(i)}\right).
\end{equation}
$\mathcal{D}_{\mathrm{TP}}$ denotes the set of correctly detected disruptive discharges, and $N_{\mathrm{TP}}$ denotes its cardinality.

In the following sections, Section~\ref{section:Multimodal Analysis} presents the experimental results and analysis of the multimodal teacher network. Section~\ref{section:Distillation Analysis} presents the experimental results of the distilled model.
Section~\ref{section:generalization} reports the generalization and scalability analyses on the independent generalization dataset and the larger-scale dataset.
Section~\ref{section:Visualization} discusses the interpretability of disruption prediction based on both the multimodal model and the distilled model.

\subsection{Analysis of the multimodal model}
\label{section:Multimodal Analysis}

% \begin{table}[t]
% \centering
% \caption{Performance comparison under model-specific optimal thresholds.}
% \label{tab:performance_comparison}
% \setlength{\tabcolsep}{3pt}
% \small
% \resizebox{\linewidth}{!}{
% \begin{tabular}{ccccccccccc}
% \toprule
% Train Mode & Test Mode & Threshold & TPR & FPR & F1 & AUC & 
% \makecell{Avg. Alarm\\(ms)} & \makecell{Inf.\\(ms)} & 
% \makecell{FLOPs\\(G)} & \makecell{Params\\(M)} \\
% % \multicolumn{11}{c}{Model-specific threshold} \\
% \midrule
% Time series & Time series & 0.32 & 91.60\% & 13.69\% & 78.57\% & 97.17\% & 1064.13 & 3.75 & 9.75 & 30.26 \\
% Video & Video & 0.94 & 83.33\% & 0.00\% & 90.90\% & 99.88\% & 14.40 & 4.10 & 21.60 & 30.02 \\
% Multimodal & Multimodal & 0.66 & 100.0\% & 2.73\% & 96.00\% & 99.62\% & 127.54 & 8.09 & 31.35 & 58.03 \\
% Multimodal & Time series & 0.53 & 95.83\% & 6.84\% & 88.46\% & 98.00\% & 776.52 & 3.75 & 9.75 & 30.26 \\
% \midrule
% \multicolumn{11}{c}{Alarm Time $>$ 10} \\
% \midrule
% Time series & Time series & 0.32 & 91.66\% & 13.69\% & 78.57\% & 97.17\% & 1064.13 & 3.75 & 9.75 & 30.26 \\
% Video & Video & 0.94 & 62.50\% & 0.00\% & 76.92\% & 99.88\% & 17.8 & 4.10 & 21.60 & 30.02 \\
% Multimodal & Multimodal & 0.66 & 100\% & 2.73\% & 96.00\% & 99.62\% & 127.54 & 8.09 & 31.35 & 58.03 \\
% Multimodal & Time series & 0.53 & 95.83\% & 6.84\% & 88.46\% & 98.00\% & 776.52 & 3.75 & 9.75 & 30.26 \\
% \bottomrule
% \end{tabular}
% }
% \end{table}

\begin{table}[t]
\centering
\caption{Performance comparison on the main test set under validation-selected thresholds.}
\label{tab:performance_comparison}
\setlength{\tabcolsep}{3pt}
\small
\resizebox{\linewidth}{!}{
\begin{tabular}{ccccccccccc}
\toprule
Train Mode & Test Mode & Threshold & TPR & FPR & F1 & AUC & 
\makecell{Avg. Alarm\\(ms)} & \makecell{Inf.\\(ms)} & 
\makecell{FLOPs\\(G)} & \makecell{Params\\(M)} \\
% \multicolumn{11}{c}{Model-specific threshold} \\
% \midrule
% Time series & Time series & 0.61 & 95.83\% & 8.21\% & 86.79\% & 97.40\% & 754.78 & 3.75 & 9.75 & 30.26 \\
% Video & Video & 0.75 & 95.83\% & 2.73\% & 93.87\% & 96.17\% & 125.95 & 4.10 & 21.60 & 30.02 \\
% Multimodal & Multimodal & 0.83 & 100.00\% & 2.73\% & 96.00\% & 99.51\% & 199.20 & 8.09 & 31.35 & 58.03 \\
% Multimodal & Time series & 0.23 & 95.83\% & 5.47\% & 90.19\% & 97.31\% & 795.65 & 3.75 & 9.75 & 30.26 \\
% \midrule
% \multicolumn{11}{c}{Alarm Time $>$ 10} \\
\midrule
Time series & Time series & 0.61 & 95.83\% & 8.21\% & 86.79\% & 96.86\% & 754.78 & 3.75 & 9.75 & 30.26 \\
Video & Video & 0.75 & 87.50\% & 2.73\% & 89.36\% & 95.80\% & 137.14 & 4.10 & 21.60 & 30.02 \\
Multimodal & Multimodal & 0.83 & 100.00\% & 2.73\% & 96.00\% & 99.00\% & 199.20 & 8.09 & 31.35 & 58.03 \\
Multimodal & Time series & 0.63 & 91.66\% & 2.73\% & 91.66\% & 97.88\% & 713.36 & 3.75 & 9.75 & 30.26 \\
\bottomrule
\end{tabular}
}
\end{table}

\begin{figure}[h]
	
	\subfigure{
		\includegraphics[width=0.48\textwidth]{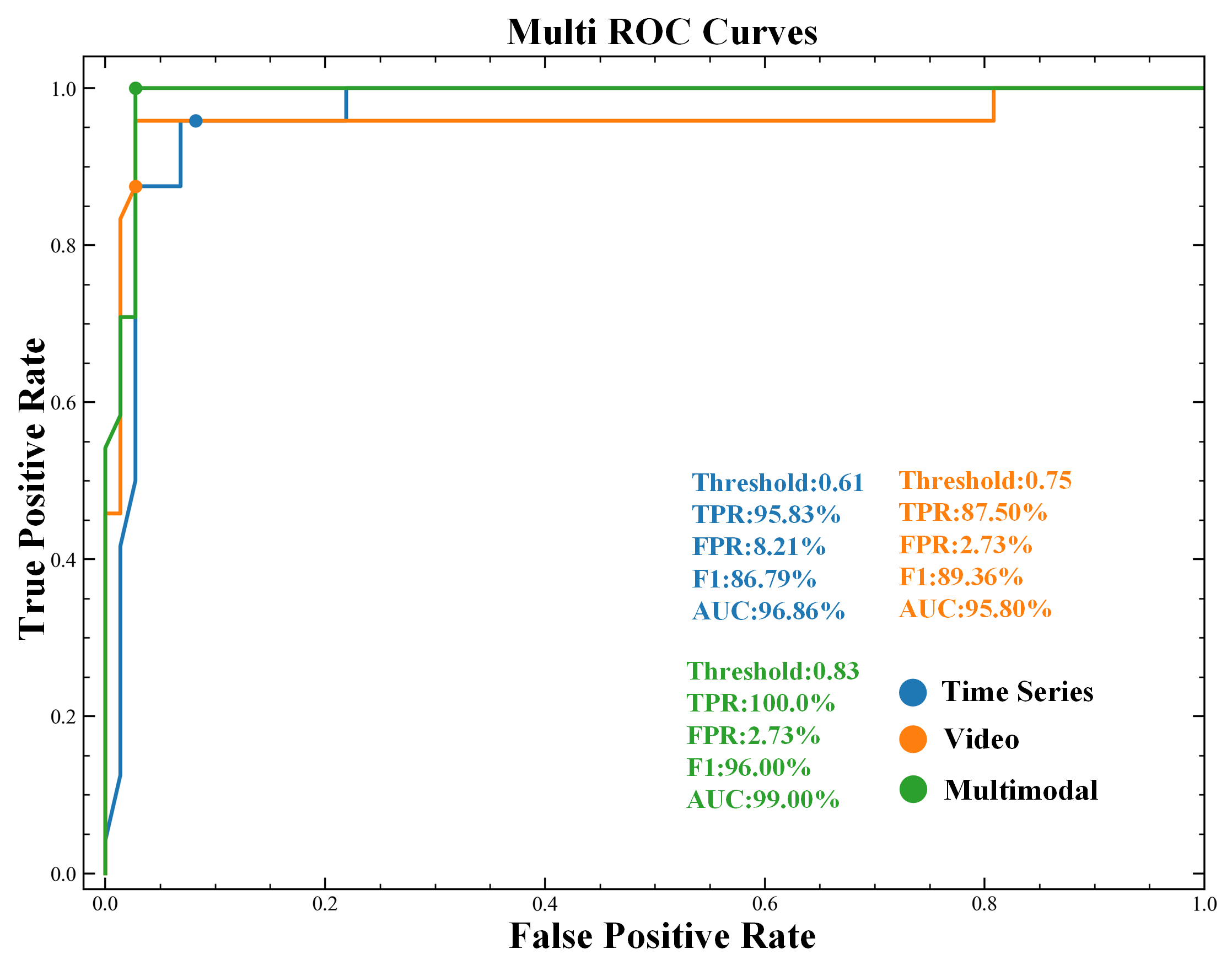}
	}
	\subfigure{
		\includegraphics[width=0.48\textwidth]{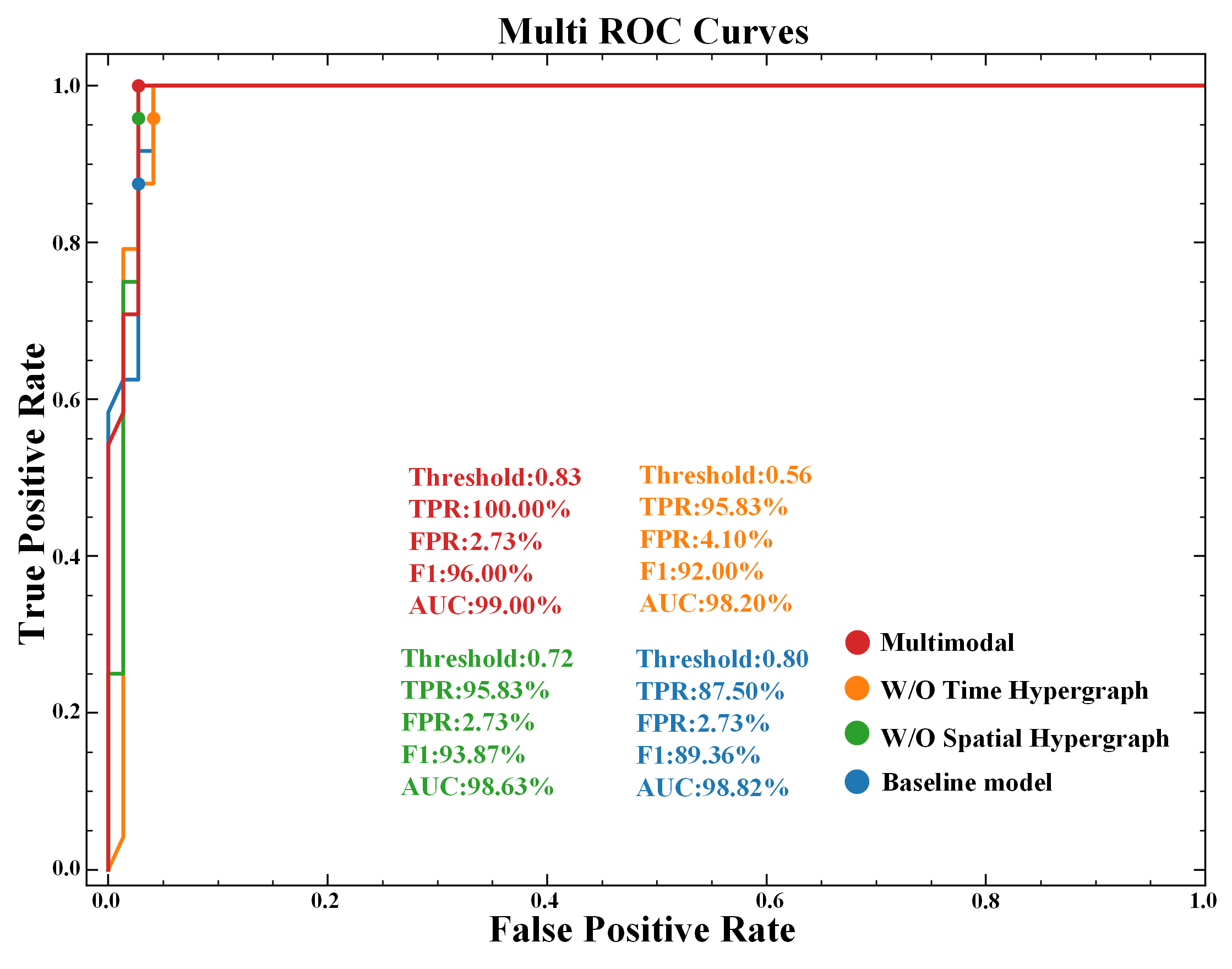}
		}
    \caption{Performance comparison of different models for EAST disruption prediction. The left panel shows the test-set ROC curves of the time-series, video, and multimodal models, with operating points determined by validation-selected thresholds. The right panel presents the ablation results of the multimodal framework, illustrating the contributions of the temporal and spatial hypergraphs to prediction performance.}  
	\label{fig:modality_ROC}
	
\end{figure}

As shown in Table~\ref{tab:performance_comparison} and the left panel of Figure~\ref{fig:modality_ROC}, this section systematically compares the time-series, video, and multimodal models for EAST disruption prediction from the perspectives of classification performance and warning capability.

\begin{figure*}[h]
    \centering
    \includegraphics[width=0.48\textwidth]{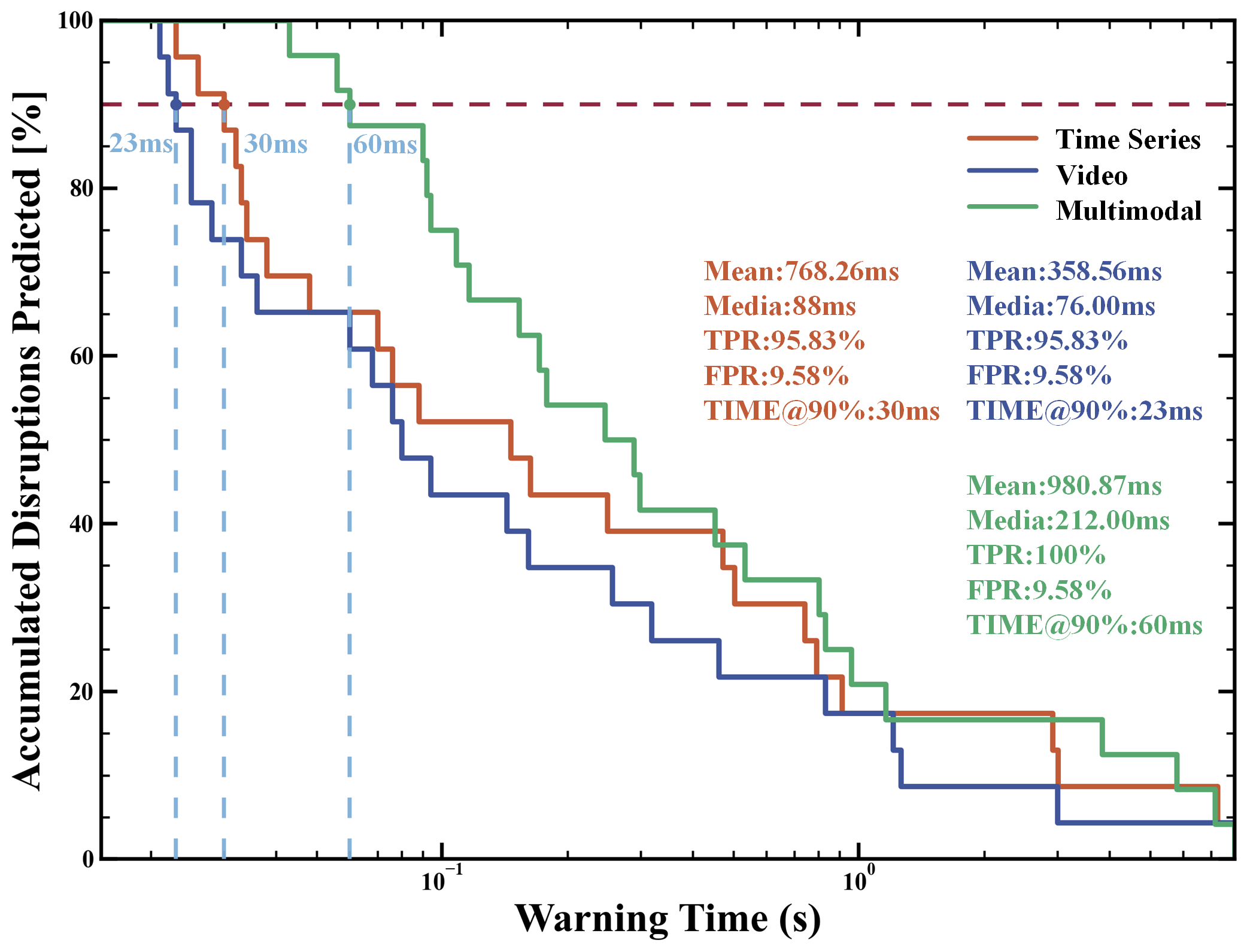}
    % \caption{Cumulative distribution of test-set warning times.}
    \caption{Cumulative distribution of warning times on the test set under a matched FPR, comparing the early-warning capability of the time-series, video, and multimodal models.}
    % A curve distributed further to the right indicates that the model can issue alarms earlier before disruption.}
    \label{fig:modality_yujin}
\end{figure*}

% 改动
Under the validation-selected thresholds, the time-series branch achieves a TPR of 95.83\%, an FPR of 8.21\%, an F1 score of 86.79\%, and an AUC of 96.86\%, while the video branch achieves a TPR of 87.50\%, an FPR of 2.73\%, an F1 score of 89.36\%, and an AUC of 95.80\%. 
Compared with the time-series branch, the video branch provides a lower FPR and a higher F1 score, indicating that visible images provide spatial discriminative information that helps suppress false alarms.
On the other hand, the time-series branch provides a longer average warning time, suggesting that time-series diagnostic signals are more sensitive to earlier-stage disruption precursors.
In contrast, the multimodal model achieves a more favorable overall trade-off, with a TPR of 100.00\%, an FPR of 2.73\%, an F1 score of 96.00\%, and an AUC of 99.00\%. 
Compared with the time-series branch, the multimodal model reduces the FPR from 8.21\% to 2.73\% and improves the F1 score from 86.79\% to 96.00\%.
Compared with the video branch, it further improves the TPR, F1 score, and AUC while maintaining the same FPR. 
These results indicate that multimodal fusion can better integrate the early temporal evolution information from diagnostic signals and the spatial abnormality information from visible images, thereby improving the overall performance of EAST disruption prediction.

To compare the early-warning capability of different models more fairly, Figure~\ref{fig:modality_yujin} presents the cumulative distribution of warning times under a fixed FPR close to 9.58\%.
It can be observed that the time-series model tends to provide a longer warning lead time, whereas the video model usually provides stronger spatial abnormality discrimination but a relatively shorter warning time. 
By integrating both types of information, the multimodal model achieves a better balance between false-alarm control and warning lead time.

The right panel of Figure~\ref{fig:modality_ROC} further presents the ablation results of the multimodal framework, where the baseline model denotes the encoder-only architecture without temporal and spatial hypergraphs.
Compared with the baseline model, the full multimodal model improves TPR by 12.50\%, F1 by 6.64\%, and AUC by 0.28\%, while keeping FPR unchanged. 
Furthermore, removing either the temporal hypergraph or the spatial hypergraph leads to performance degradation. 
These results indicate that both the temporal hypergraph and the spatial hypergraph are indispensable for achieving the final AUC of 99\%, and that the temporal hypergraph contributes more substantially to the overall performance.

\subsection{Analysis of hierarchical knowledge distillation}
\label{section:Distillation Analysis}

\begin{figure*}[t]
\centering
\includegraphics[width=\textwidth]{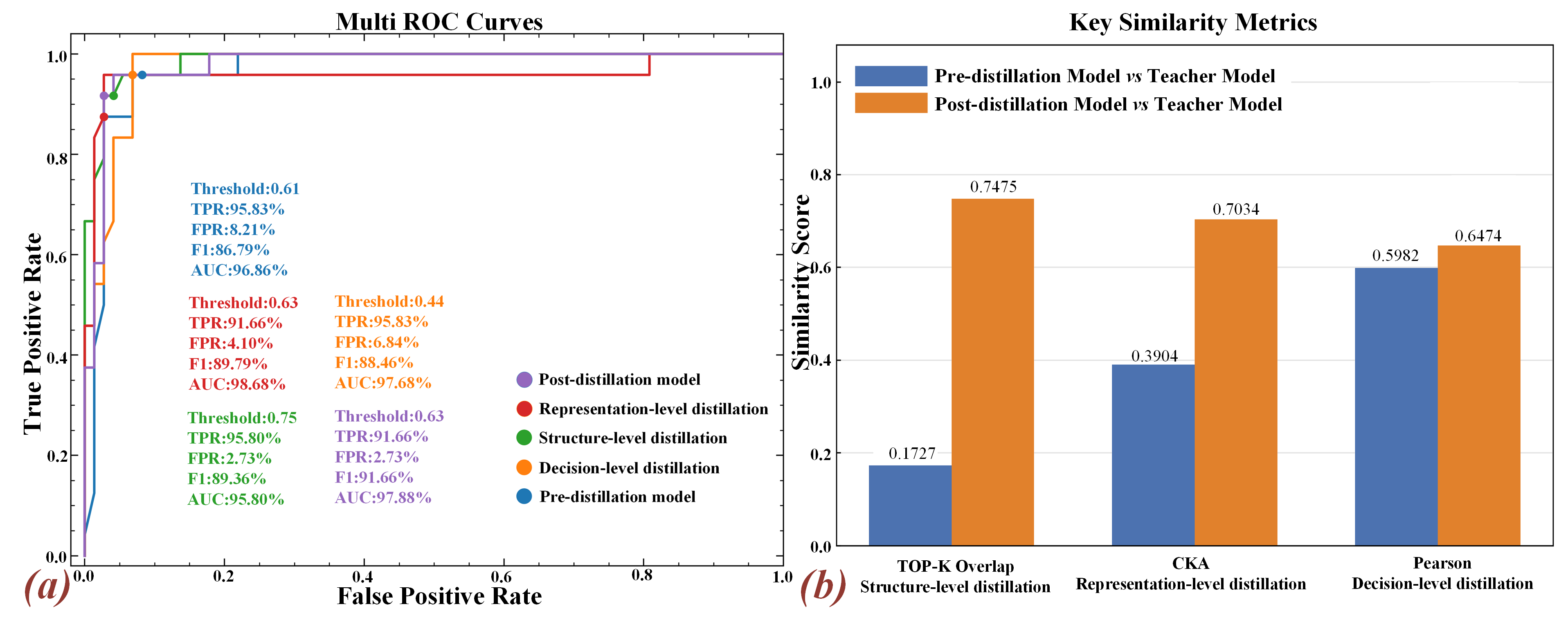}
\caption{Performance and similarity analysis of hierarchical knowledge distillation. The left panel shows the ROC curves of the pre-distillation model, the structure-level distillation model, the representation-level distillation model, the decision-level distillation model, and the post-distillation model.
The right panel compares the similarity between the student and teacher models at the structure, representation, and decision levels before and after distillation.}
\label{fig:distillation_ROC_hebing2}
\end{figure*}

As shown in the left panel of Figure~\ref{fig:distillation_ROC_hebing2} and Table~\ref{tab:performance_comparison}, the pre-distillation model is taken as the baseline.
At a threshold of 0.61, it achieves a TPR of 95.83\%, an FPR of 8.21\%, an F1 score of 86.79\%, and an AUC of 96.86\%. 
After full hierarchical distillation, the post-distillation student attains a TPR of 91.66\%, an FPR of 2.73\%, an F1 score of 91.66\%, and an AUC of 97.88\% at a threshold of 0.63. 
Compared with the pre-distillation model, the distilled model exhibits a slight reduction in TPR, while achieving a substantial decrease in FPR and improvements in both F1 score and AUC. 
These results indicate that hierarchical knowledge distillation effectively suppresses false alarms while preserving strong detection capability, thereby yielding a more favorable overall warning-performance trade-off. 
Furthermore, a comparison among the three single-level distillation variants reveals that structure-level, representation-level, and decision-level distillation transfer complementary knowledge from the teacher to the student at different levels, whereas the full hierarchical distillation achieves a more favorable overall trade-off.

In terms of inference efficiency, the multimodal model is slower because it has to process both image and time-series inputs. 
As reported in Table~\ref{tab:performance_comparison}, the multimodal model requires 8.09~ms for inference, whereas the distilled time-series student requires only 3.75~ms. 
Therefore, while retaining strong predictive performance, the distilled student improves the inference speed by about 2.16$\times$ relative to the multimodal model.
In addition, from the perspective of computational complexity and model size, the distilled time-series student maintains the same computational cost as the original time-series model, with $9.75\mathrm{G}$ FLOPs and $30.26\mathrm{M}$ parameters. Compared with the multimodal teacher, which requires $31.35\mathrm{G}$ FLOPs and $58.03\mathrm{M}$ parameters, the distilled student reduces the computational complexity and model size by 68.90\% and 47.85\%, respectively. 
These results indicate that the proposed hierarchical knowledge distillation strategy not only preserves the main discriminative capability of the multimodal teacher, but also substantially reduces the computational complexity and model size, making it more suitable for real-time disruption prediction.

To further verify whether the distilled student truly learns the teacher's knowledge, rather than merely achieving improved predictive performance, the right panel of Figure~\ref{fig:distillation_ROC_hebing2} compares the similarity between the student and teacher models at the structure, representation, and decision levels.
TOP-K Overlap measures structure-level similarity, namely the agreement between the teacher and the student in selecting the most important top-k hyperedges from the hypergraph association matrices. Its core form can be written as
\begin{equation}
\mathrm{Top\mbox{-}K\ Overlap}
=\frac{1}{M}\sum_{i=1}^{M}\frac{|A_i\cap B_i|}{|A_i\cup B_i|},
\end{equation}
where \(M\) denotes the total number of compared rows. 
For the \(i\)-th row, \(A_i\) denotes the set of top-k hyperedge indices selected by the teacher, and \(B_i\) denotes the set of top-k hyperedge indices selected by the student. The numerator \(|A_i\cap B_i|\) is the size of the intersection between the two sets, whereas the denominator \(|A_i\cup B_i|\) is the size of their union. 
TOP-K Overlap reflects the extent to which the student inherits the teacher’s sparse hypergraph association patterns. A higher value indicates that the student is more consistent with the teacher in terms of high-order structural preference.
CKA~\cite{kornblith2019similarity} measures representation-level similarity, namely the consistency between two intermediate feature spaces, and its linear form is given by
\begin{equation}
\mathrm{CKA}(X,Y)=\frac{\|X_c^\top Y_c\|_F^2}{\|X_c^\top X_c\|_F\,\|Y_c^\top Y_c\|_F},
\end{equation}
where \(X_c\) and \(Y_c\) denote the centered teacher and student feature matrices, and \(\|\cdot\|_F\) denotes the Frobenius norm. 
A larger CKA value indicates that the student is closer to the teacher in the latent representation space.
Pearson measures decision-level similarity, namely the linear consistency between the teacher and student outputs:
\begin{equation}
r(x,y)=\frac{\langle x-\bar{x},\,y-\bar{y}\rangle}{\|x-\bar{x}\|\,\|y-\bar{y}\|},
\end{equation}
where \(x\) and \(y\) denote the teacher and student output vectors, and \(\bar{x}\) and \(\bar{y}\) are their corresponding means. 
A value closer to 1 indicates that the student is more consistent with the teacher at the decision-boundary level.

As shown in the right panel of Figure~\ref{fig:distillation_ROC_hebing2}, the post-distillation student achieves substantially higher similarity scores than the pre-distillation student across all three metrics.
Specifically, the TOP-K Overlap increases from 0.1727 to 0.7475, the CKA score increases from 0.3904 to 0.7034, and the Pearson correlation increases from 0.5982 to 0.6474.
These results demonstrate that the distillation process not only improves the final predictive performance, but also makes the student significantly closer to the teacher in terms of decision boundary, intermediate representation, and high-order structural preference.
Therefore, the distilled student can be regarded as having effectively learned the multimodal knowledge of the teacher.

\subsection{Generalization and Scalability Analysis}
\label{section:generalization}

% \begin{table}[t]
% \centering
% \caption{Generalization experiment results on the EAST multimodal generalization dataset.}
% \label{tab:generalization}
% \setlength{\tabcolsep}{3pt}
% \small
% \resizebox{\linewidth}{!}{
% \begin{tabular}{ccccccccccc}
% \toprule
% Train Mode & Test Mode & Threshold & TPR & FPR & F1 & AUC & 
% \makecell{Avg. Alarm\\(ms)} & \makecell{Inf.\\(ms)} & 
% \makecell{FLOPs\\(G)} & \makecell{Params\\(M)} \\
% \midrule
% Time series & Time series & 0.32 & 87.80\% & 29.82\% & 62.88\% & 87.72\% & 1716.42 & 3.75 & 9.75 & 30.26 \\
% Video & Video & 0.94 & 85.97\% & 8.94\% & 80.57\% & 93.81\% & 175.28 & 4.10 & 21.60 & 30.02 \\
% Multimodal & Multimodal & 0.66 & 95.73\% & 10.33\% & 84.18\% & 96.21\% & 945.80 & 8.09 & 31.35 & 58.03 \\
% Multimodal & Time series & 0.53 & 86.58\% & 17.49\% & 72.08\% & 90.95\% & 1225.70 & 3.75 & 9.75 & 30.26 \\
% \midrule
% \multicolumn{11}{c}{Alarm Time $>$ 10} \\
% \midrule
% Time series & Time series & 0.32 & 85.36\% & 29.82\% & 61.67\% & 87.72\% & 1765.37 & 3.75 & 9.75 & 30.26 \\
% Video & Video & 0.94 & 75.60\% & 8.94\% & 74.47\% & 93.81\% & 198.79 & 4.10 & 21.60 & 30.02 \\
% Multimodal & Multimodal & 0.66 & 92.68\% & 10.33\% & 82.60\% & 96.21\% & 976.80& 8.09 & 31.35 & 58.03 \\
% Multimodal & Time series & 0.53 & 77.43\% & 17.49\% & 67.01\% & 90.95\% & 1369.94 & 3.75 & 9.75 & 30.26 \\
% \bottomrule
% \end{tabular}
% }
% \end{table}

\begin{table}[t]
\centering
\caption{Generalization results on the EAST multimodal generalization dataset.}
\label{tab:generalization}
\setlength{\tabcolsep}{3pt}
\small
\resizebox{\linewidth}{!}{
\begin{tabular}{ccccccccccc}
\toprule
Train Mode & Test Mode & Threshold & TPR & FPR & F1 & AUC & 
\makecell{Avg. Alarm\\(ms)} & \makecell{Inf.\\(ms)} & 
\makecell{FLOPs\\(G)} & \makecell{Params\\(M)} \\
% \multicolumn{11}{c}{Alarm Time $>$ 10} \\
\midrule
Time series & Time series & 0.61 & 78.04\% & 21.47\% & 64.00\% & 85.24\% & 1716.22 & 3.75 & 9.75 & 30.26 \\
Video & Video & 0.75 & 85.97\% & 9.74\% & 79.66\% & 93.23\% & 316.92 & 4.10 & 21.60 & 30.02 \\
Multimodal & Multimodal & 0.78 & 94.51\% & 10.93\% & 82.88\% & 94.82\% & 859.11 & 8.09 & 31.35 & 58.03 \\
Multimodal & Time series & 0.63 & 75.00\% & 10.53\% & 72.35\% & 88.06\% & 1166.9 & 3.75 & 9.75 & 30.26 \\
\bottomrule
\end{tabular}
}
\end{table}

\begin{table}[t]
\centering
\caption{Performance comparison on the larger-scale 1308-discharge dataset.}
\label{tab:performance_comparison_large_scale}
\setlength{\tabcolsep}{3pt}
\small
\resizebox{\linewidth}{!}{
\begin{tabular}{ccccccccccc}
\toprule
Train Mode & Test Mode & Threshold & TPR & FPR & F1 & AUC & 
\makecell{Avg. Alarm\\(ms)} & \makecell{Inf.\\(ms)} & 
\makecell{FLOPs\\(G)} & \makecell{Params\\(M)} \\
% \midrule
% Time series & Time series & 0.24 & 87.75\% & 13.60\% & 76.78\% & 92.65\% & 477.95& 3.75 & 9.75 & 30.26 \\
% Video & Video & 0.85 & 100\% & 6.80\% & 90.74\% & 99.11\% & 134.26 & 4.10 & 21.60 & 30.02 \\
% Multimodal & Multimodal & 0.58 & 100\% & 3.40\% & 95.14\% & 98.75\% & 188.34 & 8.09 & 31.35 & 58.03 \\
% Multimodal & Time series & 0.63 & 87.75\% & 4.76\% & 86.86\% & 96.36\% & 616.93 & 3.75 & 9.75 & 30.26 \\
% \midrule
% \multicolumn{11}{c}{Alarm Time $>$ 10} \\
\midrule
Time series & Time series & 0.24 & 81.63\% & 13.60\% & 73.39\% & 90.94\% & 513.45 & 3.75 & 9.75 & 30.26 \\
Video & Video & 0.85 & 91.83\% & 6.80\% & 86.53\% & 96.98\% & 145.88 & 4.10 & 21.60 & 30.02 \\
Multimodal & Multimodal & 0.58 & 93.38\% & 3.40\% & 92.00\% & 97.76\% & 200.36& 8.09 & 31.35 & 58.03 \\
Multimodal & Time series & 0.63 & 79.59\% & 4.76\% & 82.10\% & 91.43\% & 679.69 & 3.75 & 9.75 & 30.26 \\
\bottomrule
\end{tabular}
}
\end{table}

To further evaluate the generalization ability and scalability of the proposed framework, two additional experiments are conducted. 
First, an independent EAST multimodal generalization dataset is constructed from 668 additional discharge experiments, including 164 disruptive discharges and 504 non-disruptive discharges.
These discharges have no overlap with the 640 discharges used in the main experiments. 
Second, the original dataset and the generalization dataset are combined to form a larger-scale dataset containing 1308 discharges, and all models are retrained and evaluated using the same data-splitting strategy as described in Section~\ref{sec:Datasets}.

As shown in Table~\ref{tab:generalization}, the overall performance of all models decreases on the independent generalization dataset, indicating that unseen discharges constitute a more challenging generalization scenario. 
Nevertheless, the multimodal teacher still achieves the strongest overall performance, with a TPR of 94.51\%, an F1 score of 82.88\%, an AUC of 94.82\%, and an FPR of 10.93\%. 
This indicates that the complementarity between visible images and time-series diagnostic signals remains effective on unseen EAST discharges. 
The distilled student uses only time-series signals during inference. 
Compared with the original time-series model, it shows a slightly lower TPR, but reduces the FPR from 21.47\% to 10.53\%, improves the F1 score from 64.00\% to 72.35\%, and increases the AUC from 85.24\% to 88.06\%, while maintaining the same inference time. 
These results indicate that hierarchical knowledge distillation improves false-alarm control and overall warning performance on unseen discharges.

\begin{figure*}[t]
    \centering
    \includegraphics[width=\textwidth]{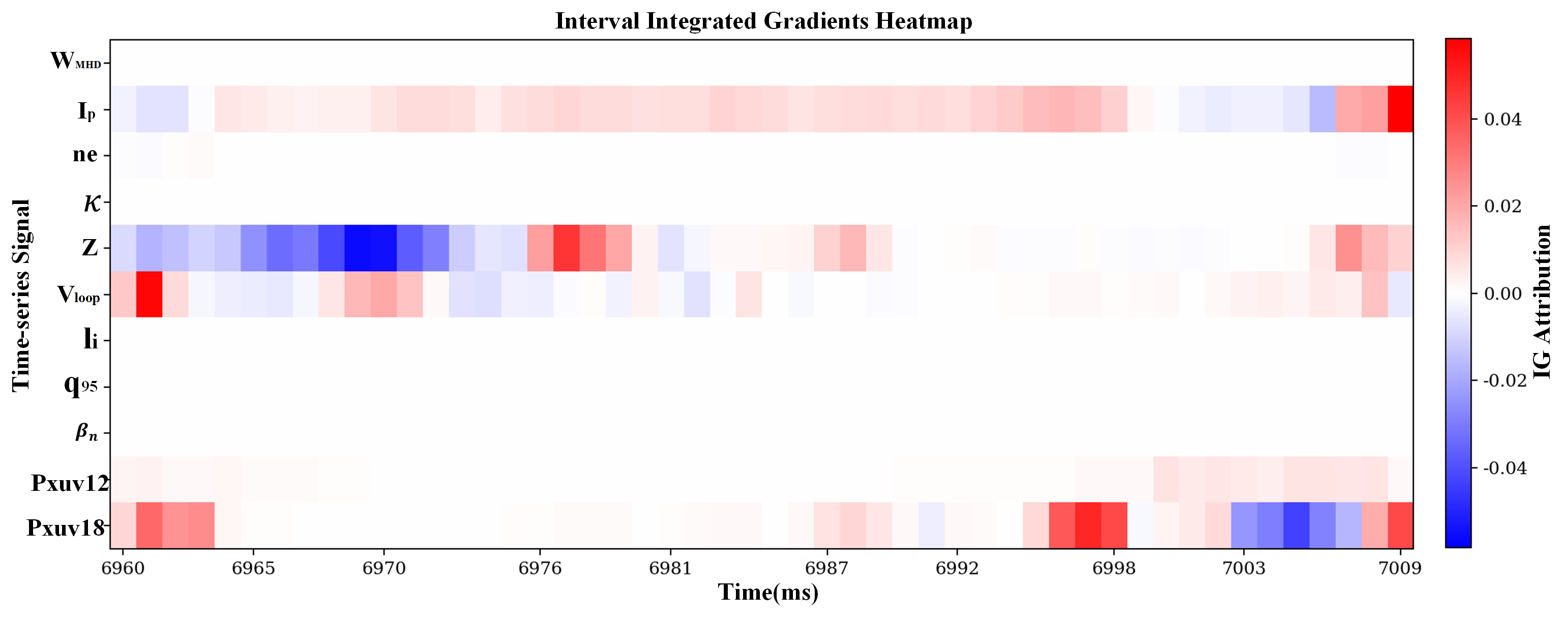}
    \caption{Integrated Gradients attribution map for input windows near the disruption time of EAST shot \#145327. The horizontal axis denotes time, and the vertical axis represents diagnostic variables. The color transition from blue to red indicates a gradually increasing contribution to the model prediction, with red regions corresponding to higher contributions.}
    \label{fig:IG_Heatmap}
\end{figure*}

\begin{figure*}[t]
    \centering
    \includegraphics[width=\textwidth]{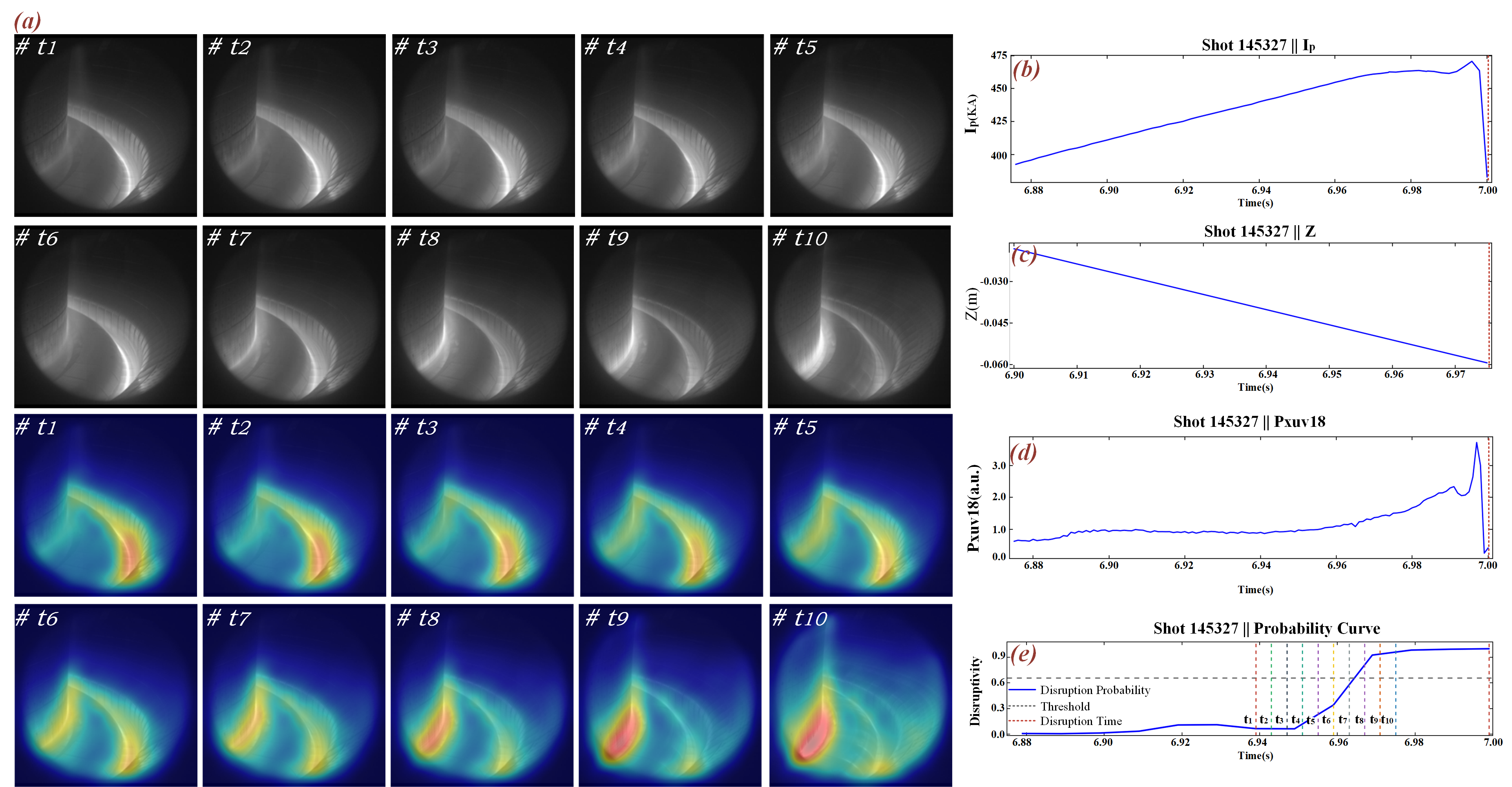}
    \caption{Visualization of multimodal model responses for EAST shot \#145327: 
    (a) original visible images and corresponding Grad-CAM activation maps; 
    (b) evolution of plasma current $I_p$; 
    (c) variation of vertical displacement $Z$; 
    (d) evolution of $P_{\mathrm{xuv}18}$, which characterizes the enhancement of impurity radiation in the plasma;
    (e) predicted disruption probability and alarm threshold.}
    \label{fig:CAM}
\end{figure*}

Furthermore, as shown in Table~\ref{tab:performance_comparison_large_scale}, on the large-scale dataset comprising 1308 discharges, the multimodal teacher model still achieves strong performance, with a TPR of 93.38\%, an FPR of 3.40\%, an F1 score of 92.00\%, and an AUC of 97.76\%. 
Compared with the original time-series model, the distilled student exhibits a slightly lower TPR, but reduces the FPR from 13.60\% to 4.76\%, while improving the F1 score from 73.39\% to 82.10\% and the AUC from 90.94\% to 91.43\%. 
It should be noted that the benefit of distillation becomes more pronounced on the larger-scale dataset.
This can be attributed to the relatively limited distributional diversity of the 640-discharge dataset, where the time-series model can already learn sufficiently discriminative representations under ground-truth supervision.
By contrast, the 1308-discharge dataset contains more diverse discharge conditions and more complex precursor evolution patterns, for which a time-series-only model may not fully capture the fine-grained discriminative variations. 
Under this setting, the multimodal teacher provides additional visual-spatial cues and smoother decision boundaries through hierarchical distillation, thereby enabling the time-series student to achieve more substantial performance gains.
These results further demonstrate that multimodal fusion and hierarchical knowledge distillation remain stable and effective as the data scale increases, thereby improving the practical applicability of efficient unimodal disruption prediction.

\subsection{Visualization}
\label{section:Visualization}

\begin{figure*}[t]
    \centering
    \includegraphics[width=\textwidth]{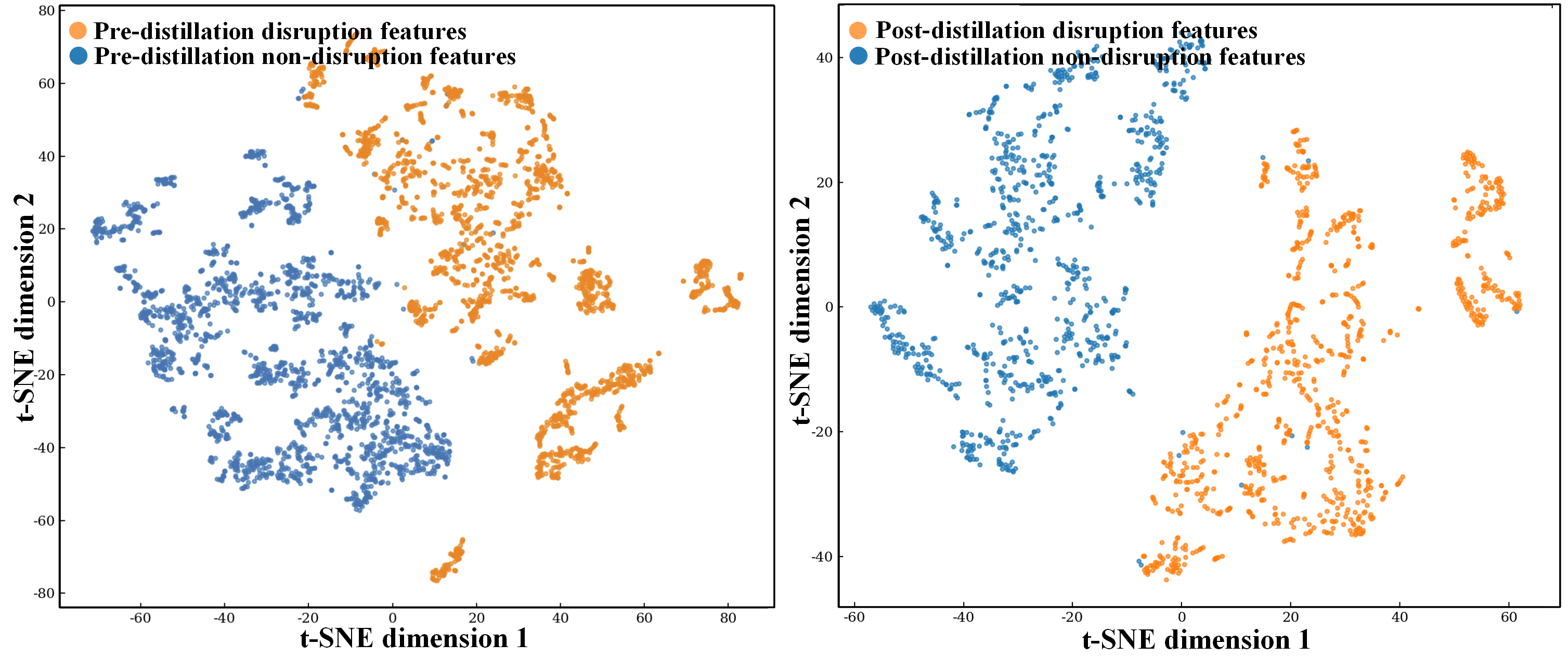}
    \caption{t-SNE visualization of disruptive and non-disruptive student features. 
    The left panel shows the feature distribution before distillation, while the right panel shows the feature distribution after distillation, where larger inter-class separation and a clearer class boundary can be observed.}
    \label{fig:tsne}
\end{figure*}

As shown in Figure~\ref{fig:vis_dataset}, EAST shot~\#145327 is mainly characterized by impurity-radiation-induced disruption.
% Taking EAST shot \#145327 as an example, this discharge is mainly characterized as a disruption triggered by enhanced impurity radiation. 
As shown in Figure~\ref{fig:IG_Heatmap}, Integrated Gradients~\cite{sundararajan2017axiomatic} is applied to the input window close to the disruption time.
The attribution results show that, at approximately $6.998\mathrm{s}$, the model mainly focuses on radiation-related features associated with $P_{\mathrm{xuv}18}$, indicating that this disruption is closely related to impurity-radiation enhancement.
Figure~\ref{fig:CAM}(c) further supports this conclusion, the $Z$ coordinate of the plasma current centroid decreases continuously, while Figure~\ref{fig:CAM}(a) shows that a distinct radiation band gradually forms in the lower region of the visible-light image. 
These observations indicate that the plasma undergoes both vertical displacement evolution and radiation-structure enhancement before the disruption.
As the radiation signal continues to increase, the predicted disruption probability rises continuously from $t_1$ to $t_{10}$ and exceeds the alarm threshold at $t_8$, issuing an early warning approximately $16\mathrm{ms}$ before the actual disruption.
The corresponding Grad-CAM activation maps~\cite{selvaraju2017grad} in the lower row of Figure~\ref{fig:CAM}(a) show that the model begins to exhibit a stronger response to the plasma and high-radiation regions from $t_8$ onward. 
These results demonstrate good consistency among the Integrated Gradients attribution maps, the Grad-CAM activation maps, and the temporal evolution of the diagnostic signals.
This indicates that the multimodal model can capture physically meaningful precursor information related to impurity radiation and spatial-structure evolution, and therefore exhibits favorable physical interpretability.

To further examine the effect of distillation on the feature representation of the student model, Figure~\ref{fig:tsne} visualizes the disruptive and non-disruptive features before and after distillation using t-SNE. 
It can be observed that the distilled features exhibit a larger inter-cluster distance and a clearer class boundary.
Combined with the similarity analysis shown in the right panel of Figure~\ref{fig:distillation_ROC_hebing2}, these observations further suggest that the distilled student improves its final predictive performance by inheriting the discriminative information encoded in the multimodal teacher network.

\section{Conclusions}
\label{sec:Conclusions}

In this paper, we propose a hierarchical multi-to-single-modal knowledge distillation framework for EAST disruption prediction, aiming to reduce the high inference cost of multimodal models while preserving their superior discriminative capability.
Based on a synchronized EAST multimodal dataset composed of visible images and time-series diagnostic signals, a multimodal teacher is constructed during training to learn richer disruption precursor representations through Transformer-based modality encoders and a prototype-guided spatiotemporal hypergraph module. 
On this basis, a multi-level distillation strategy is further developed to progressively transfer the teacher’s knowledge at the graph-structure, representation, and decision levels to a time-series-based student, thereby establishing a multimodal-training but unimodal-inference framework.

Experimental results show that, using validation-selected thresholds with the 10 ms valid-warning criterion, the multimodal teacher achieves strong performance and obtains the highest TPR and F1 score among the compared models, supporting the complementarity between visible images and time-series diagnostic signals for disruption prediction in EAST.
Additional experiments on the independent generalization dataset and the larger-scale dataset further verify the robustness and scalability of the proposed framework.
Although some performance degradation is observed on unseen discharges, the multimodal teacher model still maintains strong overall warning capability.
The distilled student usually achieves a substantial reduction in FPR and improvements in F1 score and AUC without increasing inference cost, thereby improving overall warning performance while maintaining efficient inference.
Moreover, the distilled student achieves an approximately $2.16\times$ inference speedup over the multimodal teacher and reduces the FLOPs and parameter count by approximately $68.90\%$ and $47.85\%$, respectively.
Similarity and visualization analyses further show that the distilled student effectively inherits the teacher's discriminative knowledge and that the model responses are consistent with physically meaningful disruption precursors, indicating favorable physical interpretability.

Overall, this study verifies the feasibility of combining multimodal-enhanced representation learning with efficient unimodal inference for disruption prediction in EAST, and provides an effective framework for reducing inference cost while preserving strong predictive capability. Future work may further extend this framework to richer diagnostic modalities and cross-device generalization scenarios.

\section*{Acknowledgments}
This work was supported by the National Natural Science Foundation of China under Grant Nos. U24A20342 and 12475233, and the National Key Research and Development Program of China under Grant No. 2025YFE0201800. 
The authors would like to acknowledge the support and contributions from the rest of the EAST team, the High-performance Computing Platform of Anhui University for providing computing resources. 
\section*{References}

\bibliographystyle{unsrt}
\bibliography{references}

@article{hender2007chapter,
  title   = {Chapter 3: {MHD} stability, operational limits and disruptions},
  author  = {Hender, T. C. and Wesley, J. C. and Bialek, J. and Bondeson, A. and Boozer, A. H. and Buttery, R. J. and Garofalo, A. and Goodman, T. P. and Granetz, R. S. and Gribov, Y. and others},
  journal = {Nuclear Fusion},
  volume  = {47},
  number  = {6},
  pages   = {S128--S202},
  year    = {2007}
}

@article{boozer2012theory,
  title={Theory of tokamak disruptions},
  author={Boozer, Allen H},
  journal={Physics of plasmas},
  volume={19},
  number={5},
  year={2012},
  publisher={AIP Publishing}
}

@inproceedings{paszke2019pytorch,
  title={PyTorch: An Imperative Style, High-Performance Deep Learning Library},
  author={Paszke, Adam and Gross, Sam and Massa, Francisco and Lerer, Adam and Bradbury, James and Chanan, Gregory and Killeen, Trevor and Lin, Zeming and Gimelshein, Natalia and Antiga, Luca and Desmaison, Alban and Kopf, Andreas and Yang, Edward and DeVito, Zachary and Raison, Martin and Tejani, Alykhan and Chilamkurthy, Sasank and Steiner, Benoit and Fang, Lu and Bai, Junjie and Chintala, Soumith},
  booktitle={Advances in Neural Information Processing Systems},
  volume={32},
  pages={8024--8035},
  year={2019}
}

@article{zhang2019graph,
  title={Graph convolutional networks: a comprehensive review},
  author={Zhang, Si and Tong, Hanghang and Xu, Jiejun and Maciejewski, Ross},
  journal={Computational Social Networks},
  volume={6},
  number={1},
  pages={1--23},
  year={2019},
  publisher={Springer}
}

@inproceedings{loshchilov2019decoupled,
  title={Decoupled Weight Decay Regularization},
  author={Loshchilov, Ilya and Hutter, Frank},
  booktitle={International Conference on Learning Representations},
  year={2019}
}

@article{sugihara2007disruption,
  title   = {Disruption scenarios, their mitigation and operation window in {ITER}},
  author  = {Sugihara, M. and Shimada, M. and Fujieda, H. and Gribov, Yu and Ioki, K. and Kawano, Y. and Khayrutdinov, R. and Lukash, Victor and Ohmori, J.},
  journal = {Nuclear Fusion},
  volume  = {47},
  number  = {4},
  pages   = {337--352},
  year    = {2007}
}

@article{kates2019predicting,
  title   = {Predicting disruptive instabilities in controlled fusion plasmas through deep learning},
  author  = {Kates-Harbeck, Julian and Svyatkovskiy, Alexey and Tang, William},
  journal = {Nature},
  volume  = {568},
  number  = {7753},
  pages   = {526--531},
  year    = {2019}
}

@article{murari2020transfer,
  title   = {On the transfer of adaptive predictors between different devices for both mitigation and prevention of disruptions},
  author  = {Murari, A. and Rossi, R. and Peluso, E. and Lungaroni, M. and Gaudio, P. and Gelfusa, M. and Ratt{\'a}, G. and Vega, J. and {JET Contributors} and {ASDEX Upgrade Team}},
  journal = {Nuclear Fusion},
  volume  = {60},
  number  = {5},
  pages   = {056003},
  year    = {2020}
}

@article{zheng2020disruption,
  title   = {Disruption predictor based on neural network and anomaly detection on {J-TEXT}},
  author  = {Zheng, W. and Wu, Q. Q. and Zhang, M. and Chen, Z. Y. and Shang, Y. X. and Fan, J. N. and Pan, Y. and {J-TEXT Team}},
  journal = {Plasma Physics and Controlled Fusion},
  volume  = {62},
  number  = {4},
  pages   = {045012},
  year    = {2020}
}

@article{rea2019real,
  title={A real-time machine learning-based disruption predictor in {DIII-D}},
  author={Rea, Christina and Montes, KJ and Erickson, KG and Granetz, RS and Tinguely, RA},
  journal={Nuclear Fusion},
  volume={59},
  number={9},
  pages={096016},
  year={2019},
  publisher={IOP Publishing}
}

@article{hu2021real,
  title={Real-time prediction of high-density {EAST} disruptions using random forest},
  author={Hu, WH and Rea, Cristina and Yuan, QP and Erickson, KG and Chen, DL and Shen, Biao and Huang, Yao and Xiao, JY and Chen, JJ and Duan, YM and others},
  journal={Nuclear Fusion},
  volume={61},
  number={6},
  pages={066034},
  year={2021},
  publisher={IOP Publishing}
}

@article{vega2013results,
  title   = {Results of the {JET} real-time disruption predictor in the {ITER}-like wall campaigns},
  author  = {Vega, Jes{\'u}s and Dormido-Canto, Sebasti{\'a}n and L{\'o}pez, Juan M. and Murari, Andrea and Ram{\'\i}rez, Jes{\'u}s M. and Moreno, Ra{\'u}l and Ruiz, Mariano and Alves, Diogo and Felton, Robert and {JET-EFDA Contributors} and others},
  journal = {Fusion Engineering and Design},
  volume  = {88},
  number  = {6-8},
  pages   = {1228--1231},
  year    = {2013}
}

@article{aymerich2022disruption,
  title   = {Disruption prediction at {JET} through deep convolutional neural networks using spatiotemporal information from plasma profiles},
  author  = {Aymerich, Enrico and Sias, Giuliana and Pisano, Fabio and Cannas, Barbara and Carcangiu, Sara and Sozzi, Carlo and Stuart, Chris and Carvalho, P. J. and Fanni, Alessandra and {JET Contributors}},
  journal = {Nuclear Fusion},
  volume  = {62},
  number  = {6},
  pages   = {066005},
  year    = {2022}
}

@article{pau2018first,
  title   = {A first analysis of {JET} plasma profile-based indicators for disruption prediction and avoidance},
  author  = {Pau, Alessandro and Fanni, A. and Cannas, B. and Carcangiu, S. and Pisano, Gianluca and Sias, G. and Sparapani, P. and Baruzzo, M. and Murari, A. and Rimini, F. and others},
  journal = {IEEE Transactions on Plasma Science},
  volume  = {46},
  number  = {7},
  pages   = {2691--2698},
  year    = {2018}
}

@article{zhu2021hybrid,
  title   = {Hybrid deep-learning architecture for general disruption prediction across multiple tokamaks},
  author  = {Zhu, J. X. and Rea, Cristina and Montes, Kevin and Granetz, R. S. and Sweeney, Ryan and Tinguely, Roy Alexander},
  journal = {Nuclear Fusion},
  volume  = {61},
  number  = {2},
  pages   = {026007},
  year    = {2021}
}

@article{aledda2015improvements,
  title   = {Improvements in disruption prediction at {ASDEX Upgrade}},
  author  = {Aledda, Raffaele and Cannas, Barbara and Fanni, Alessandra and Pau, Alessandro and Sias, Giuliana and {ASDEX Upgrade Team} and others},
  journal = {Fusion Engineering and Design},
  volume  = {96},
  pages   = {698--702},
  year    = {2015}
}

@article{rea2018disruption,
  title   = {Disruption prediction investigations using machine learning tools on {DIII-D} and {Alcator C-Mod}},
  author  = {Rea, Cristina and Granetz, R. S. and Montes, K. and Tinguely, Roy Alexander and Eidietis, N. and Hanson, Jeremy M. and Sammuli, B.},
  journal = {Plasma Physics and Controlled Fusion},
  volume  = {60},
  number  = {8},
  pages   = {084004},
  year    = {2018}
}

@article{yoshino2003neural,
  title   = {Neural-net disruption predictor in {JT-60U}},
  author  = {Yoshino, R.},
  journal = {Nuclear Fusion},
  volume  = {43},
  number  = {12},
  pages   = {1771--1786},
  year    = {2003}
}

@article{guo2021disruption,
  title   = {Disruption prediction using a full convolutional neural network on {EAST}},
  author  = {Guo, B. H. and Shen, B. and Chen, D. L. and Rea, C. and Granetz, R. S. and Huang, Y. and Zeng, L. and Zhang, H. and Qian, J. P. and Sun, Y. W. and others},
  journal = {Plasma Physics and Controlled Fusion},
  volume  = {63},
  number  = {2},
  pages   = {025008},
  year    = {2021}
}

@article{guo2023disruption,
  title   = {Disruption prediction on {EAST} with different wall conditions based on a multi-scale deep hybrid neural network},
  author  = {Guo, B. H. and Chen, D. L. and Rea, C. and Wu, M. Q. and Shen, B. and Granetz, R. S. and Zhang, Z. C. and Huang, Y. and Duan, Y. M. and Zeng, L. and others},
  journal = {Nuclear Fusion},
  volume  = {63},
  number  = {9},
  pages   = {094001},
  year    = {2023}
}

@article{zheng2018hybrid,
  title   = {Hybrid neural network for density limit disruption prediction and avoidance on {J-TEXT} tokamak},
  author  = {Zheng, W. and Hu, F. R. and Zhang, M. and Chen, Z. Y. and Zhao, X. Q. and Wang, X. L. and Shi, P. and Zhang, X. L. and Zhang, X. Q. and Zhou, Y. N. and others},
  journal = {Nuclear Fusion},
  volume  = {58},
  number  = {5},
  pages   = {056016},
  year    = {2018}
}

@article{shen2023idp,
  title   = {{IDP-PGFE}: an interpretable disruption predictor based on physics-guided feature extraction},
  author  = {Shen, Chengshuo and Zheng, Wei and Ding, Yonghua and Ai, Xinkun and Xue, Fengming and Zhong, Yu and Wang, Nengchao and Gao, Li and Chen, Zhipeng and Yang, Zhoujun and others},
  journal = {Nuclear Fusion},
  volume  = {63},
  number  = {4},
  pages   = {046024},
  year    = {2023}
}

@article{zhong2021disruption,
  title   = {Disruption prediction and model analysis using {LightGBM} on {J-TEXT} and {HL-2A}},
  author  = {Zhong, Y. and Zheng, W. and Chen, Z. Y. and Xia, F. and Yu, L. M. and Wu, Q. Q. and Ai, X. K. and Shen, C. S. and Yang, Z. Y. and Yan, W. and others},
  journal = {Plasma Physics and Controlled Fusion},
  volume  = {63},
  number  = {7},
  pages   = {075008},
  year    = {2021}
}

@article{yang2020disruption,
  title   = {A disruption predictor based on a 1.5-dimensional convolutional neural network in {HL-2A}},
  author  = {Yang, Zongyu and Xia, Fan and Song, Xianming and Gao, Zhe and Huang, Yao and Wang, Shuo},
  journal = {Nuclear Fusion},
  volume  = {60},
  number  = {1},
  pages   = {016017},
  year    = {2020}
}

@article{yang2021depth,
  title   = {In-depth research on the interpretable disruption predictor in {HL-2A}},
  author  = {Yang, Zongyu and Xia, Fan and Song, Xianming and Gao, Zhe and Wang, Shuo and Dong, Yunbo},
  journal = {Nuclear Fusion},
  volume  = {61},
  number  = {12},
  pages   = {126042},
  year    = {2021}
}

@article{yang2025implementing,
  title   = {Implementing deep learning-based disruption prediction in a drifting data environment of new tokamak: {HL-3}},
  author  = {Yang, Zongyu and Zhong, Wulyu and Xia, Fan and Gao, Zhe and Zhu, Xiaobo and Li, Jiyuan and Hu, Liwen and Xu, Zhaohe and Li, Da and Zheng, Guohui and others},
  journal = {Nuclear Fusion},
  volume  = {65},
  number  = {2},
  pages   = {026030},
  year    = {2025}
}

@article{kwon2021tokamak,
  title   = {Tokamak visible image sequence recognition using nonlocal spatio-temporal {CNN} for attention needed area localization},
  author  = {Kwon, Giil and Wi, Hanmin and Hong, Jaesic},
  journal = {Fusion Engineering and Design},
  volume  = {168},
  pages   = {112375},
  year    = {2021}
}

@inproceedings{joshi2023analysis,
  title     = {Analysis of Different Inference Implementations for Deep Learning Model on {ADITYA-U} Tokamak},
  author    = {Joshi, Ramesh and Ghosh, Joydeep and Kalani, Nilesh and Tanna, R. L.},
  booktitle = {International Conference on Soft Computing for Problem-Solving},
  pages     = {145--155},
  year      = {2023},
  publisher = {Springer}
}

@incollection{chandrasekaran2021data,
  title     = {Data-Driven-Based Disruption Prediction in {GOLEM} Tokamak with Missing Values},
  author    = {Chandrasekaran, Jayakumar and Madhawa, Surendar and Sangeetha, J.},
  booktitle = {Intelligent Systems, Technologies and Applications: Proceedings of Sixth {ISTA} 2020, India},
  pages     = {129--149},
  year      = {2021},
  publisher = {Springer}
}

@article{scarpari2024st40,
  title   = {{ST40} electromagnetic predictive studies supported by machine learning applied to experimental database},
  author  = {Scarpari, M. and Minucci, S. and Sias, G. and Lombroni, R. and Buxton, P. F. and Romanelli, M. and Calabr{\`o}, G.},
  journal = {Scientific Reports},
  volume  = {14},
  number  = {1},
  pages   = {27074},
  year    = {2024}
}

@article{poels2025plasma,
  title   = {Plasma state monitoring and disruption characterization using multimodal {VAEs}},
  author  = {Poels, Yoeri and Pau, Alessandro and Donner, Christian and Romanelli, Giulio and Sauter, Olivier and Venturini, Cristina and Menkovski, Vlado and {TCV Team} and {WPTE Team}},
  journal = {Nuclear Fusion},
  volume  = {65},
  number  = {9},
  pages   = {096012},
  year    = {2025}
}

@article{liu2025interpretable,
  title   = {An interpretable disruption predictor on {EAST} using improved {XGBoost} and {SHAP}},
  author  = {Liu, D. M. and Zhu, X. L. and Jiang, Y. S. and Wang, S. and Shu, S. B. and Shen, B. and Guo, B. H. and Liu, L. C.},
  journal = {Nuclear Fusion},
  volume  = {65},
  number  = {8},
  pages   = {086035},
  year    = {2025}
}

@article{lee2025real,
  title   = {Real-time data-driven disruption prediction and its mitigation of {MA}-plasma experiments in {KSTAR} with a lower carbon divertor},
  author  = {Lee, Jeongwon and Hahn, Sang-hee and Han, Hyunsun and Kim, Jayhyun and Juhn, June-Woo and Bak, Jun-Gyo and Song, Jae-in and Nam, Yong Un},
  journal = {Nuclear Fusion},
  volume  = {65},
  number  = {5},
  pages   = {056040},
  year    = {2025}
}

@article{lee2025machine,
  title   = {Machine learning based disruption prediction using long short-term memory in {KSTAR}},
  author  = {Lee, Jeongwon and Kim, Jayhyun and Kim, Jinsu and Hahn, Sang-hee and Han, Hyunsun and Shin, Giwook and Na, Yong-Su and Nam, Yong Un},
  journal = {Nuclear Fusion},
  volume  = {65},
  number  = {8},
  pages   = {086017},
  year    = {2025}
}

@article{shen2024cross,
  title   = {Cross-tokamak disruption prediction based on domain adaptation},
  author  = {Shen, Chengshuo and Zheng, Wei and Guo, Bihao and Ding, Yonghua and Chen, Dalong and Ai, Xinkun and Xue, Fengming and Zhong, Yu and Wang, Nengchao and Shen, Biao and others},
  journal = {Nuclear Fusion},
  volume  = {64},
  number  = {6},
  pages   = {066036},
  year    = {2024}
}

@article{ratta2021phad,
  title   = {{PHAD}: a phase-oriented disruption prediction strategy for avoidance, prevention, and mitigation in {JET}},
  author  = {Ratt{\'a}, Giuseppe A. and Vega, Jes{\'u}s and Murari, Andrea and Gadariya, Dhaval and {JET Contributors}},
  journal = {Nuclear Fusion},
  volume  = {61},
  number  = {11},
  pages   = {116055},
  year    = {2021}
}

@article{spolladore2023detection,
  title   = {Detection of {MARFEs} using visible cameras for disruption prevention},
  author  = {Spolladore, L. and Rossi, R. and Wyss, I. and Gaudio, P. and Murari, A. and Gelfusa, M. and {JET Contributors}},
  journal = {Fusion Engineering and Design},
  volume  = {190},
  pages   = {113507},
  year    = {2023}
}

@article{chen2026vi,
  title   = {{Vi-DP}: low-latency video-based disruption prediction with multi-{FOV} fusion in {EAST}},
  author  = {Chen, M. W. and Yang, K. J. and Zhu, J. J. and Wang, Q. J. and Cheng, F. and Chen, D. L. and Gong, X. Z. and Zhang, B. and Zhang, X. H. and Duan, Y. M. and others},
  journal = {Nuclear Fusion},
  volume  = {66},
  number  = {1},
  pages   = {016002},
  year    = {2026}
}

@article{kim2024disruption,
  title   = {Disruption prediction and analysis through multimodal deep learning in {KSTAR}},
  author  = {Kim, Jinsu and Lee, Jeongwon and Seo, Jaemin and Lee, Yeongsun and Na, Yong-Su},
  journal = {Fusion Engineering and Design},
  volume  = {200},
  pages   = {114204},
  year    = {2024}
}

@article{kim2024enhancing,
  title   = {Enhancing disruption prediction through {B}ayesian neural network in {KSTAR}},
  author  = {Kim, Jinsu and Lee, Jeongwon and Seo, Jaemin and Ghim, Young-Chul and Lee, Yeongsun and Na, Yong-Su},
  journal = {Plasma Physics and Controlled Fusion},
  volume  = {66},
  number  = {7},
  pages   = {075001},
  year    = {2024}
}

@inproceedings{kornblith2019similarity,
  title     = {Similarity of neural network representations revisited},
  author    = {Kornblith, Simon and Norouzi, Mohammad and Lee, Honglak and Hinton, Geoffrey},
  booktitle = {Proceedings of the 36th International Conference on Machine Learning},
  pages     = {3519--3529},
  year      = {2019},
  publisher = {PMLR}
}

@inproceedings{sundararajan2017axiomatic,
  title     = {Axiomatic attribution for deep networks},
  author    = {Sundararajan, Mukund and Taly, Ankur and Yan, Qiqi},
  booktitle = {Proceedings of the 34th International Conference on Machine Learning},
  pages     = {3319--3328},
  year      = {2017},
  publisher = {PMLR}
}

@inproceedings{selvaraju2017grad,
  title     = {{Grad-CAM}: visual explanations from deep networks via gradient-based localization},
  author    = {Selvaraju, Ramprasaath R. and Cogswell, Michael and Das, Abhishek and Vedantam, Ramakrishna and Parikh, Devi and Batra, Dhruv},
  booktitle = {Proceedings of the IEEE International Conference on Computer Vision},
  pages     = {618--626},
  year      = {2017}
}

\end{document}